%% file: main.tex
\documentclass[10pt,twocolumn,letterpaper]{article}

\usepackage[pagenumbers]{iccv} % To force page numbers, e.g. for an arXiv version


\usepackage[table,xcdraw,dvipsnames]{xcolor}
\definecolor{iccvblue}{rgb}{0.21,0.49,0.74}
\usepackage[pagebackref,breaklinks,colorlinks,allcolors=iccvblue]{hyperref}
\PassOptionsToPackage{table}{xcolor} 
\usepackage{graphicx}
\usepackage{amsmath}
\usepackage{amssymb}
\usepackage{booktabs}

\usepackage{multirow}
\usepackage{hhline}
\usepackage{arydshln}
\usepackage{comment}
\usepackage{enumitem}
\usepackage{caption}
\usepackage{makecell}

\usepackage{algorithm}
\usepackage{algpseudocode}

\definecolor{Gray}{gray}{0.9}

\def\paperID{*****} % *** Enter the Paper ID here
\def\confName{ICCV}
\def\confYear{2025}

\title{DynFaceRestore: Balancing Fidelity and Quality in Diffusion-Guided Blind Face Restoration with Dynamic Blur-Level Mapping and Guidance}

\author{
Huu-Phu Do$^{1}$\thanks{Equal contribution}, \qquad Yu-Wei Chen$^{1}$\footnotemark[1],  \qquad  Yi-Cheng Liao$^{1}$,\qquad  Chi-Wei Hsiao$^{2}$, \\ Han-Yang Wang$^{2}$,   \qquad Wei-Chen Chiu$^{1}$, \qquad Ching-Chun Huang$^{1}$\thanks{Corresponding author, Ching-Chun Huang (chingchun@nycu.edu.tw)}\\
\\
$^{1}$National Yang Ming Chiao Tung University, Taiwan \qquad  $^{2}$MediaTek Inc., Taiwan \\
}

\newcommand{\CCH}[1]{{\color{black}#1}\normalfont} %black

\begin{document}
\maketitle
\input{sec/0_abstract}    
\input{sec/1_intro}
\input{sec/2_related_works}
\input{sec/3_preliminary}
\input{sec/4_method}
\input{sec/5_experiments}

\input{sec/6_conclusion}
\appendix\input{sec/X_suppl}

{
    \small
    \bibliographystyle{ieeenat_fullname}
    \bibliography{main}
}

\end{document}

%% file: sec/0_abstract.tex
\CCH{
\begin{abstract}
Blind Face Restoration aims to recover high-fidelity, detail-rich facial images from unknown degraded inputs, presenting significant challenges in preserving both identity and detail. Pre-trained diffusion models have been increasingly used as image priors to generate fine details. Still, existing methods often use fixed diffusion sampling timesteps and a global guidance scale, assuming uniform degradation. This limitation and potentially imperfect degradation kernel estimation frequently lead to under- or over-diffusion, resulting in an imbalance between fidelity and quality. We propose DynFaceRestore, a novel blind face restoration approach that learns to map any blindly degraded input to Gaussian blurry images. By leveraging these blurry images and their respective Gaussian kernels, we dynamically select the starting timesteps for each blurry image and apply closed-form guidance during the diffusion sampling process to maintain fidelity. Additionally, we introduce a dynamic guidance scaling adjuster that modulates the guidance strength across local regions, enhancing detail generation in complex areas while preserving structural fidelity in contours. This strategy effectively balances the trade-off between fidelity and quality. DynFaceRestore achieves state-of-the-art performance in both quantitative and qualitative evaluations, demonstrating robustness and effectiveness in blind face restoration. Project page at \href{https://nycu-acm.github.io/DynFaceRestore/}{Link}
\end{abstract}
}

%Blind Face Restoration involves recovering high-fidelity, detail-rich facial images from unknown degraded inputs, presenting challenges in preserving identity and detail. To ensure high-quality details in the restored images, pre-trained diffusion models as high-quality image priors for generating details have gained attention. However, existing methods uniformly treat degradation severity, applying fixed sampling timesteps and global-wise guidance scale, resulting in under-diffuse or over-diffuse and losing the balance between fidelity and quality. In this paper, we propose DynFaceRestore, a blind face restoration approach thatfirst maps degraded inputs to Gaussian blurry images based on degradation severity, allowing the dynamic selection of starting timesteps and enabling closed-form guidance in following the sampling process to maintain fidelity. In addition, we use a dynamic guidance scaling adjuster that modulates guidance strength across local regions, enhancing detail generation in complex areas while preserving structural fidelity in contours, further balancing the trade-off between fidelity and quality. DynFaceRestore achieves state-of-the-art results in both quantitative and qualitative evaluations, demonstrating robustness and effectiveness.

%% file: sec/1_intro.tex
\section{Introduction}
\label{sec:intro}

\begin{figure}[htp]
    \centering
    \includegraphics[width=0.9\linewidth]{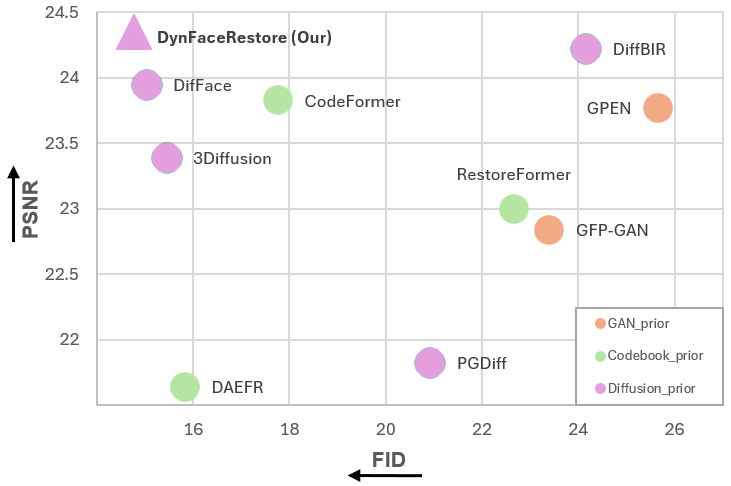}
    \caption{Blind face restoration demands both high fidelity and rich detail. Compared to other SOTA methods that leverage GAN priors, codebook priors, or diffusion priors, our proposed method, DynFaceRestore (denoted as an asterisk), demonstrates superior image fidelity (PSNR$\uparrow$) and quality (FID$\downarrow$) on the CelebA-Test.}
    \label{fig:psnr-fid}
\end{figure}

\begin{figure}[htp]
    \centering
    \includegraphics[scale=0.6]{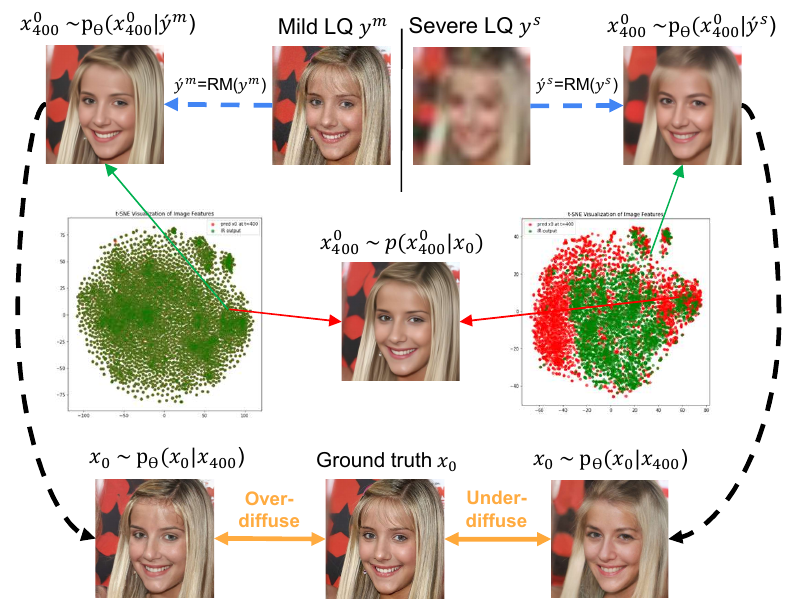}
    \caption{Using DifFace \cite{yue2024difface} as an example, let $RM$ represent DifFace's restoration model and $p_{\Theta}$ the diffusion model. In the t-SNE plot, green and red points denote the features of $x_{400}^{0}$ sampled from $p_{\Theta}(x_{400}^{0}|\acute{y})$ and $p_{\Theta}(x_{400}^{0}|x_{0})$, respectively. Here, $\acute{y}$, the LQ image restored by $RM$, is used for diffusion guidance. DifFace initiates the diffusion process at a fixed $t = 400$, resulting in under-diffusion (right) for severely degraded LQ images and just enough diffusion for mildly degraded LQ images (left). This underscores the importance of selecting an appropriate starting step.}
   % Under/Over-diffuse issue. Using DifFace \cite{yue2024difface} as an example, where $RM$ denotes DifFace's restoration model and $p_{\Theta}$ denotes the diffusion model. Green and red points in the t-SNE plot represent the features of $x_{400}^{0}$ sampled from $p_{\Theta}(x_{400}^{0}|\acute{y})$ and $p_{\Theta}(x_{400}^{0}|x_{0})$, respectively. Using $\acute{y}$, the LQ image restored by RM for guidance, DifFace starts the diffusion process at fixed $t= 400$, leading to under-diffusion for severe LQ  and over-diffusion for leads mild LQ. This implies the importance of selecting the diffusion starting step. } %to under-diffusion, while mild LQ causes over-diffusion due to a domain gap.
    \label{fig:under-over-diffuse}
\end{figure}

Blind face restoration (BFR) involves recovering high-fidelity, detail-rich facial images from degraded inputs with unknown kernels \cite{wang2022survey,li2023survey}. The main difficulty lies in enhancing facial details while preserving the individual's identity, as shown in \cref{fig:psnr-fid}. Although GAN-based approaches \cite{chen2021psfrgan, yang2021gpen, kumar2022gfpgan} are effective at adding fine details, they often introduce artifacts. Recently, diffusion models (DM) \cite{ho2020denoising} have demonstrated great generative capabilities to create more realistic details. However, DMs can still compromise fidelity in the restored faces.

% Previous methods \cite{yang2024pgdiff,yue2024difface} have demonstrated that using pre-trained DM with high-quality (HQ) image priors can effectively assist in generating realistic facial details; however, these methods have suffered from several issues. 
% First, these methods apply a restoration model to remove degradation and restore structure before using DM to enhance details. However, these methods treat all degradation severity of LQ inputs as same and use a fixed starting timestep to sample, leading to the under-diffuse or over-diffuse after using DM to add details, as illustrated in Fig.\ref{fig:under-over-diffuse}. In addition, the distribution of the output of the restoration model is hard to define in a closed-form function, leading to the difficulty of using the output of the restoration model to guide the sampling process.
% Second, the region such as hair and wrinkle, should leverage more DM power to generate details for perceptual quality. In contrast, the regions such as facial contours, should utilize more guidance to keep fidelity. However, previous methods \cite{fei2023gdp,yang2024pgdiff} using guidance to guide DM typically apply a fixed scale factor to all pixels and lack a local region view of the image, which leads to sub-optimal perceptual quality or loose fidelity.

Previous approaches \cite{yang2024pgdiff,yue2024difface, fei2023gdp} have effectively demonstrated the use of pre-trained diffusion models (DM) with high-quality (HQ) image priors to generate realistic facial details. Typically, \cite{yang2024pgdiff,yue2024difface}  apply DM for fine detail enhancement after mitigating degradation and recovering the structural information of low-quality (LQ) images using a restoration model. Meanwhile, \cite{chung2023blinddps, fei2023gdp} aims to maintain image structures by guiding the DM through the measurement model. Despite these advancements, these methods still face challenges in maintaining fidelity. Below, we outline possible reasons for this limitation.

First, these methods assume that all LQ inputs exhibit uniform degradation severity and thus use a fixed starting timestep for diffusion sampling in image restoration. This approach often results in under- or over-diffusion when the DM adds details, as illustrated in \cref{fig:under-over-diffuse}. Second, in blind settings, imperfect kernel estimation and arbitrary kernel forms frequently cause kernel mismatch issues, which misguide the diffusion sampling process. Third, high-frequency regions like hair and wrinkles benefit from stronger DM influence to enhance perceptual quality. In contrast, low-frequency regions, like facial contours, require greater guidance from the observed image to preserve fidelity. However, existing methods \cite{chung2023blinddps,fei2023gdp, yang2024pgdiff} that incorporate guidance into DM generally apply a uniform scale factor across all pixels, overlooking region-specific variations. 
% DiffBIR \cite{diffbir} uses the Sobel operator on the LR image to directly set the local guidance level.
This lack of localized adjustment can lead to suboptimal perceptual quality or fidelity loss, hindering the balance between detail enhancement and structural integrity.

In this paper, we introduce DynFaceRestore, which leverages a pre-trained diffusion model to enhance fine details while preserving fidelity through multiple Dynamic Blur-Level Mapping and Guidance. First, we design a discriminative restoration network that maps a blindly degraded input image into multiple Gaussian-blurred versions with estimated blur levels, effectively transforming unknown, variable degradation into a Gaussian form. Since perfect kernel estimation is often impractical, using multiple Gaussian-blurred images as multi-guidance for DM sampling helps address kernel mismatch issues. Second, the estimated blur level determines a dynamic starting timestep for each Gaussian-blurred image, ensuring optimal guidance insertion during diffusion and effectively addressing under- and over-diffusion challenges. Third, to enable localized adjustment, we implement DPS-based guidance \cite{chung2023blinddps, chung2022dps} and introduce a novel dynamic guidance scale adjuster that adapts the guidance scale region-wise, resulting in high-fidelity, detail-rich outputs, as illustrated in \cref{fig:psnr-fid}. Our main contributions are summarized as follows:

%In this paper, we propose DynFaceRestore, which leverages a pre-trained DM to enhance fine details while balancing fidelity using Dynamic Multi-Blur-Level Mapping Dynamic Blur-level Mapping (DBM). First, we design a discriminative restoration network to map a blindly degraded input image into multiple Gaussian blurred versions, with the estimated blur levels. This helps to convert the unknown kernel with varying degradation levels to a Gaussian form. Meanwhile, since a perfect kernel estimation is impractical, the usage of multiple Gaussian blurred versions as multi-guidance for DM sampling help to reveal the kernel mismatching issue effectively. Second, the estimated blur level suggests a dynamic starting timestep to insert the guidance for each Gaussian blurred image, addressing under- and over-diffuse issue effectively. Third, to address the lack of localized adjustment, we employ DPS-based guidance \cite{chung2023blinddps,chung2022dps} and introduce a dynamic guidance scale adjuster to adapt the guidance scale region-wise, achieving results that are both detail-rich and high fidelity, as illustrated in Fig.\ref{fig:psnr-fid}. Our main contributions can be summarized as follows:

%straightforward approach maps unknown degradation levels to a Gaussian blur, with the blur level controlled by the standard deviation (std). Specifically, our restoration model maps inputs with varying levels of degradation severity to Gaussian blurry images with corresponding blur levels. These blur levels are then used to determine the starting timestep.

\begin{itemize}%[leftmargin=10pt, parsep=-3pt]
% \item A discriminative restoration network that maps a blindly degraded image to multiple Gaussian blur images, enabling multiple closed-form guidance for guiding diffusion sampling process to preserve high fidelity.
\item We introduce Dynamic Blur-Level Mapping and Guidance to address kernel mismatch issues effectively, using multi-level Gaussian-blurred images as guidance.
\item Leveraging the estimated blur level from the restoration network output, we dynamically determine the optimal starting step in the diffusion process to prevent under- and over-diffusion.
\item We construct a dynamic guidance scale adjuster to tailor the guidance strength across regions, harnessing the powerful influence of DM to improve perceptual quality while maintaining fidelity.
\item Our proposed method, DynFaceRestore, outperforms state-of-the-art approaches in both quantitative and qualitative evaluations, demonstrating its superiority in Blind Face Restoration.
\end{itemize}

%% file: sec/2_related_works.tex
\begin{figure*}[htp]
    \centering
    \includegraphics[scale=0.5]{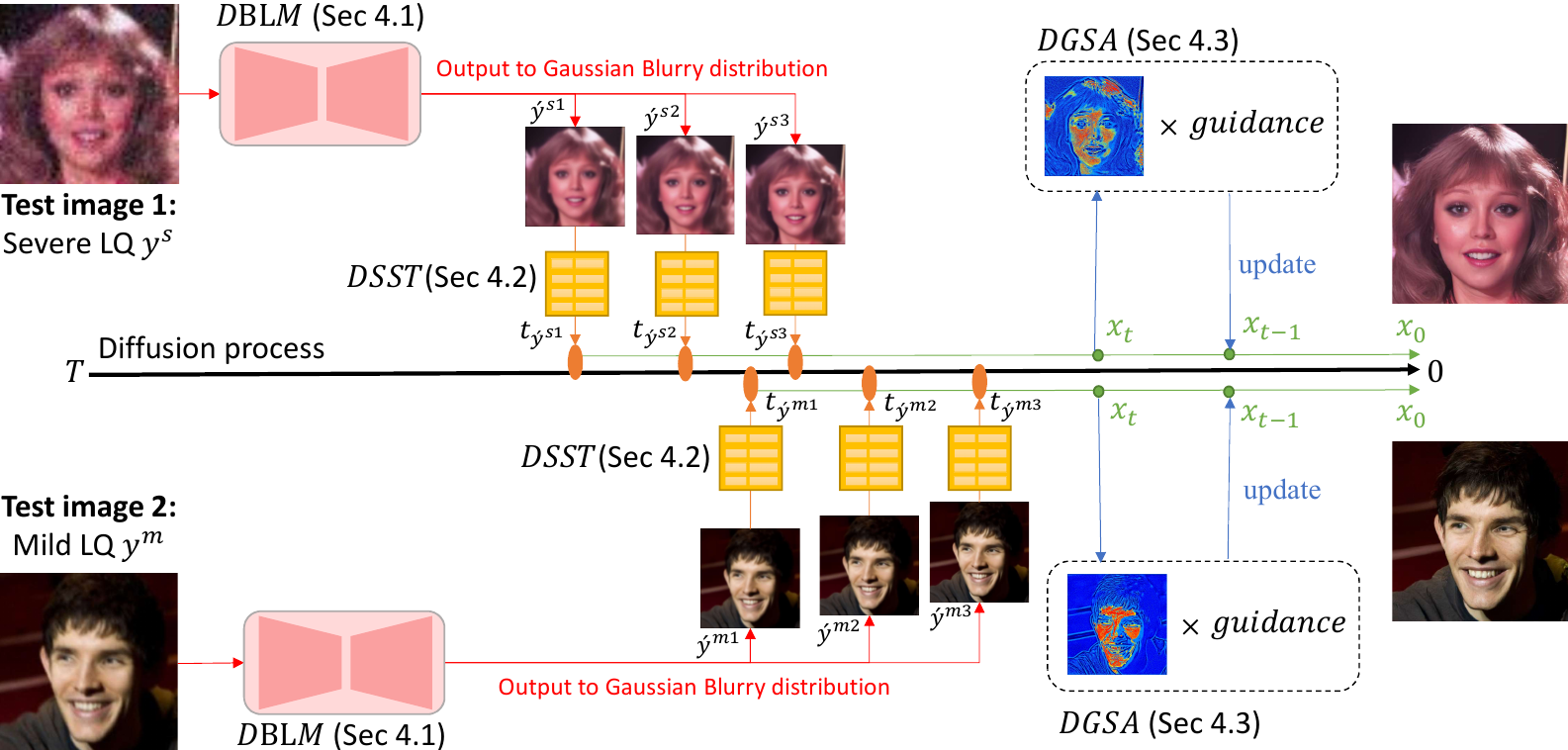}
    \caption{Overview of our proposed DynFaceRestore framework, which consists of three key components: DBLM, DSST, and DGSA (defined in \cref{sec:method}). The upper and lower sections illustrate two independent restoration scenarios with inputs degraded to varying levels. DBLM generates multiple Gaussian-blurred images based on the degradation level of the unknown degraded input. Then, given these blur levels, DSST identifies the optimal starting step for each Gaussian-blurred image via a predefined lookup table, providing sampling guidance to avoid under- or over-diffusion. Lastly, the trained network, DGSA, locally adjusts the guidance scale used in the pre-trained diffusion process, enabling DynFaceRestore to achieve an optimal balance between fidelity and quality.}
    \label{fig:dynamicface}
\end{figure*}

\section{Related Works}
\label{sec:related}

\subsection{Blind Face Restoration (BFR) with Priors}
Previous BFR methods have utilized various priors to enhance restoration accuracy. Geometric prior methods \cite{chen2021psfrgan, chen2018fsrnet, shen2018deep} rely on structural data, such as landmarks and parsing maps, but often struggle when degraded inputs lack structural details. Reference prior methods \cite{tsai2024dualassociatedencoderface, wang2022restoreformer, zhou2022codeformer,gu2022vqfr} use high-quality (HQ) reference images or codebooks. For example, VQ-GAN \cite{esser2021vqgan} provides detailed facial features; however, its fixed codebook size can limit the diversity and richness of restored faces. GAN-based approaches, such as PSFRGAN \cite{chen2021psfrgan}, incorporate facial parsing maps for style control, while models like GFP-GAN \cite{kumar2022gfpgan} and GPEN \cite{yang2021gpen} focus on restoring facial fidelity by embedding pre-trained face models \cite{karras2020analyzing, karras2019style}. Although GAN-based methods enhance face restoration quality, they often introduce artifacts.

\CCH{
Recently, diffusion-based approaches for BFR have achieved SOTA performance by either training a DM from scratch \cite{suin2024diffuseandrestore, miao2024waveface, qiu2023diffbfr, chen2024blind, 3Diffusion} or guiding a pre-trained DM \cite{dr2, yang2024pgdiff, yue2024difface, diffbir} to provide priors. For example, WaveFace \cite{miao2024waveface} employs a conditional DM to generate high-frequency components, while 3Diffusion \cite{3Diffusion} integrates 3D facial structure into the noise estimation process. However, these methods typically require extensive training resources.  

Other approaches leverage pre-trained diffusion models for restoration. DR2 \cite{dr2} removes degradation using a pre-trained DM and enhances resolution via a super-resolution module. DiffBIR \cite{diffbir} employs a restoration module to remove degradation and regenerates details using a conditional pre-trained DM with ControlNet. DifFace \cite{yue2024difface} maps a LQ input to an intermediate diffusion sampling step, then applies the standard DM backward process for restoration without extra guidance. In contrast, PGDiff \cite{yang2024pgdiff} adopts classifier guidance to constrain the denoising process.

However, many of these DM-based methods overlook local degradation severity and rely on a fixed diffusion starting step, leading to suboptimal restoration. Recently, \cite{diffbir} addressed this issue by applying the Sobel operator on the LR image to set local guidance levels and introduce varying priors from DM. Yet, it struggles to capture intricate textures, as it focuses on local intensity without understanding global structures. In contrast, our proposed method provides ``step-wise and region-specific guidance'', effectively adapting to both local and global features to preserve complex textures, as further discussed in \cref{sec:Dynamic Guidance Scaling Estimator}.
}

%In contrast, our DynFaceRestore dynamically adjusts degradation removal and contour restoration, mapping LQ inputs to Gaussian blurry levels corresponding to degradation severity, thus preserving fidelity and enhancing detail. 

\subsection{Guidance on unconditional DM}
Methods such as DDRM \cite{kawar2022ddrm}, DDNM \cite{wang2022ddnm}, BlindDPS \cite{chung2023blinddps}, Fast-diffusion EM \cite{laroche2024fast}, and GDP \cite{fei2023gdp} avoid the need for restoration training by leveraging a pre-trained diffusion model. These approaches modify the intermediate output at each diffusion step to align with the degraded input (LQ image) as guidance. The alignment relies on a degradation model, represented by either a fixed linear matrix \cite{wang2022ddnm, kawar2022ddrm}, a kernel-based DM \cite{chung2023blinddps}, or a learnable degradation model \cite{fei2023gdp}. However, these methods generally assume a simple or linear degradation process, creating a significant domain gap when applied to complex real-world degradations. Additionally, they overlook kernel mismatch issues during diffusion, where the estimated kernel often deviates from the true degradation.

%In contrast, our DynFaceRestore first maps unknown degraded LQ images to  Gaussian Blur distribution, enabling our approach to handle real-world complex degradation inputs.

%% file: sec/3_preliminary.tex
\section{Preliminary and Framework Overview}
\label{sec:preliminary}

\subsection{Diffusion Posterior Sampling (DPS)}
\label{sec:dps}

Consider the following observation model:
\begin{equation}
\begin{split}
y=\mathcal{A}(x)+n,
\end{split}
\label{eq.inverse_problem}
\end{equation}
where $y$ is the measurement degraded from its HQ image counterpart $x$, $\mathcal{A}$ models the degradation process, and $n$ is Gaussian noise. Based on Bayes' rule, our goal is to sample or recover $x$ from the posterior distribution $p(x|y)\propto P(y|x)P(x)$. Here, the likelihood $P(y|x)$ is defined according to \cref{eq.inverse_problem}. When using a pre-trained DM to model the HQ image prior, the diffusion sampling process with $y$ as guidance at timestep $t$ can be expressed as 
\begin{equation}
\begin{split}
x_{t-1}=x_{t-1}'-s\nabla_{x_{t}}\log p_{t}(y|x_{t}),\\
\end{split}
\label{eq.ddpm_backward_guidance}
\end{equation}
where $s$ is the guidance scale, $x_{t-1}'$ is sampled from $p(x_{t-1}|x_{t})$, and $p_{t}(y|x_{t})$ is likelihood.

Diffusion Posterior Sampling \cite{chung2022dps} builds upon a pre-trained DM from the Denoising Diffusion Probabilistic Model (DDPM) \cite{ho2020denoising} and approximates the gradient of the likelihood $\nabla_{x_{t}}\log p_{t}(y|x_{t})$ as $\nabla_{x_{t}}\log p_{t}(y|\hat{x}_{t}^{0})$, where
\begin{equation}
\hat{x}_{t}^{0}=\frac{1}{\sqrt{\bar{\alpha_{t}}}}(x_{t}-\sqrt{1-\bar{\alpha}_{t}}\epsilon_{\theta}).
\label{eq.ddpm_backward}
\end{equation}
Here, $\epsilon_{\theta}$ is a learnable network used to define a diffusion model and 
${\bar{\alpha_{t}}}$ represents the pre-defined DDPM coefficients \cite{ho2020denoising}. Based on \cref{eq.ddpm_backward} and using $\epsilon_{\theta}$, we can estimate the HQ image $\hat{x}_{t}^{0}$ from the diffusion sample $x_t$. Since $\hat{x}_{t}^{0}\cong {x}$, we can utilize the observation model in \cref{eq.inverse_problem} to replace the sample guidance in \cref{eq.ddpm_backward_guidance} as follows:   
\begin{equation}
\begin{split}
\nabla_{x_{t}}\log p_{t}(y|x_{t}) \approx  \nabla_{x_t} \| y - \mathcal{A}(x_t^0) \|_2^2.
\end{split}
\label{eq.dps}
\end{equation}

Note that $\mathcal{A}$ should be given in advance. If $\mathcal{A}$ is unknown or imperfectly estimated \cite{chung2023blinddps}, an estimator or refiner for $\mathcal{A}$ is necessary. Below, we proposed modifications to DPS and the way to refine $\mathcal{A}$ during diffusion sampling.   

%BlindDPS \cite{chung2023blinddps} suggests that using guidance $\nabla_{\mathcal{A}_{t}}\log p_{t}(y|x_{t},\mathcal{A}_{t})$ to guide an additional diffusion model for $\mathcal{A}$ to estimate and refine $\mathcal{A}$ is feasible.

\subsection{DynFaceRestore Sampling}
\label{sec:ours_guidance}
Rather than directly tackling the challenging BFR task, our DynFaceRestore method reframes it as a Gaussian deblurring problem by transforming the blindly degraded input image ${y}$ into a Gaussian-blurred version $\acute{y}$ and applying a modified DPS-based guidance for restoration. This allows us to represent $\mathcal{A}(x_t^0)$ as $k_{t}\otimes x_{t}^{0}$, where $k$ is a Gaussian blur kernel with standard deviation $std$. Consequently, our guided diffusion sampling process can be reformulated as follows:
\begin{equation}
\begin{split}
x_{t-1}=x_{t-1}'-s\nabla_{x_{t}}\| \acute{y} - k_{t}\otimes x_{t}^{0} \|_2^2.
\end{split}
\label{eq.ddpm_backward_guidance_kernel}
\end{equation}
where $x_{t-1}'$ sampled from $p(x_{t-1}|x_{t})$ is defined as follows:
\begin{equation}
\begin{split}
x_{t-1}'=\frac{1}{\sqrt{\alpha_{t}}}(x_{t}-\frac{\beta_{t}}{\sqrt{1-\bar{\alpha}_{t}}}\epsilon_{\theta})+\sigma_{t}\epsilon,\space\epsilon\sim N(0,1),
\end{split}
\label{eq.ddpm_backward_guidance_kernel}
\end{equation}
which are derived from diffusion model. However, perfect kernel estimation is often impractical; to reduce kernel mismatch, we also refine the $\mathcal{A}$ during diffusion sampling. Since our $\mathcal{A}$ is strictly a Gaussian blur kernel, we only need to update the estimated $std$ by using the following equation. 
\begin{equation}
\begin{split}
std_{t-1}=std_{t}-s\nabla_{std_{t}}\| \acute{y} - k_{t}\otimes x_{t}^{0} \|_2^2.
\end{split}
\label{eq:backward_kernel_guidance}
\end{equation}
% Specifically, when updating $x_{t-1}$ using \cref{eq.ddpm_backward_guidance_kernel}, we keep $std_{t}$ fixed; conversely, when updating $std_{t-1}$ using \cref{eq:backward_kernel_guidance}, we utilize the updated $x_{t}^{0}$.

% However, to address the kernel mismatch problem and ensure optimal guidance, as mentioned in Section 1,  our framework employs multi-step guidance across distinct timesteps tailored to different blur levels.

% (1) a \textit{Dynamic Blur-Level Mapping} (DBLM), which transforms blindly degraded low-quality (LQ) inputs into multiple Gaussian-blurred versions with blur levels adapted to the degradation level of the input; 
% (2) a \textit{Dynamic Starting Step Lookup Table} (DSST), which identifies the optimal starting timestep to insert guidance into DM for each Gaussian-blurred image; and 
% (3) a \textit{Dynamic Guidance Scaling Adjuster} (DGSA), which directs the diffusion sampling process in a region-wise manner.

Additionally, as shown in \cref{fig:dynamicface}, the DynFaceRestore framework enhances DPS in the following four key aspects.

% Fig.\ref{fig:dynamicface} illustrates the overall of our framework, and  Algorithm \ref{alg:inference} introduce the inference steps:
\begin{enumerate}
\item The DBLM (\cref{sec:Dynamic Blur-Level Mapping}) first transforms a blindly degraded LQ input $y$ into the corresponding Gaussian-blurred version $\acute{y}$, with blur levels adapted to the input's degradation level, enabling the representation of $\mathcal{A}(x_t^0)$ as $k_{t}\otimes x_{t}^{0}$.
% By formulating $\mathcal{A}(x_t^0)$ as $k_{t}\otimes x_{t}^{0}$, where $k$ is a Gaussian blur kernel with standard deviation $std$ estimated from SE (Sec.\ref{sec:se}) and $x_{t}^{0}$ is the HQ prediction at timestep $t$, we guide the diffusion model sampling process as follows:
% \begin{equation}
% \begin{split}
% x_{t-1}=x_{t-1}-s\nabla_{x_{t}}\| y - k_{t}\otimes x_{t}^{0} \|_2^2
% \end{split}
% \label{eq.ddpm_backward_guidance_kernel}
% \end{equation}
% However, perfect kernel estimation is often impractical, leading to kernel mismatch. Due to $k$ is always a Gaussian blur kernel, we refine $k$ through $std$ instead of the whole kernel across timesteps as follows:
% \begin{equation}
% \begin{split}
% std_{t-1}=std_{t-1}-s\nabla_{std_{t}}\log p_{t}(y|x_{t},std_{t})
% \end{split}
% \label{eq:backward_kernel_guidance}
% \end{equation}
\item Then, through DSST table (\cref{sec:Dynamic Starting Step}), the optimal starting timestep $t_{start}$ for diffusion is identified by using the estimated kernel (i.e., $\hat{std^*}$ in \cref{sec:se}) as query. A right starting timestep would mitigate the effects of under- and over-diffusion. Next, $\acute{y}$, $\hat{std^*}$, and $t_{start}$ are used to initialize our DPS-based sampling process. 
\item Instead of using a global scale $s$, during the sampling process, the DGSA (\cref{sec:Dynamic Guidance Scaling Estimator}) dynamic adjusts guidance scale region-wise, balancing the restoration result's quality and fidelity. 
\item Finally, we extend our framework to incorporate multiple guidance for the BFR task, as detailed in \cref{sec:Multiple Mapping}.
\end{enumerate}

To better understand our work, we also outline the inference steps in \cref{alg:inference} for the reference.

\input{algorithm/inference}

%% file: algorithm/inference.tex
\begin{algorithm}[t]
\small
  \caption{Inference}
  \label{alg:inference}
  \begin{algorithmic}[1]
    \Require
      $y$: Unknown degraded LQ input; 
    \Ensure
      $x_{0}$: HQ sampled image;    
    \State $\acute{y},\hat{std^*}=DBLM(y), SE(y)$;
    \State compute Gaussian blur kernel $k$ using $\hat{std^*}$
    \State $t_{start}=DSST(\hat{std^*})$;
    \State $x_{t_{start}}=\sqrt{\bar{\alpha}_{t_{start}}}\acute{y}+\sqrt{1-\bar{\alpha}_{t_{start}}}\epsilon, \epsilon\sim N(0,1)$;
    \State $std_{t_{start}}=\hat{std^*}$;
    \For{$t=t_{start}\cdots 1$}
      \State $x_{t}^{0}=\frac{1}{\sqrt{\bar{\alpha_{t}}}}x_{t}-\sqrt{\frac{1-\bar{\alpha_{t}}}{\bar{\alpha}_{t}}}\epsilon_{\theta}$;
      \State $x_{t-1}'=\frac{1}{\sqrt{\alpha_{t}}}(x_{t}-\frac{\beta_{t}}{\sqrt{1-\bar{\alpha}_{t}}}\epsilon_{\theta})+\sigma_{t}\epsilon,\space\epsilon\sim N(0,1)$;
      % \State $std_{t-1}'=std_{t}$;
      \State compute Gaussian blur kernel $k_{t}$ using $std_{t}$;
%       \State $gd_{x_{t}}\equiv 
% \nabla_{x_{t}}\left\| \acute{y}-k_{t}\otimes x_{t}^{0} \right\|^{2}$;
%       \State $gd_{k_{t}}\equiv 
% \nabla_{k_{t}}\left\| \acute{y}-k_{t}\otimes x_{t}^{0} \right\|^{2}$;
      \State $A_t=DGSA(\acute{y},x_{t}^{0},t)$;
      % \If{$t>t_{fine}$}
      %   \State $x_{t-1}=x_{t-1}-\sqrt{\bar{\alpha}_{t}}\times \nabla_{x_{t}}\left\| \acute{y}-k_{t}\otimes x_{t}^{0} \right\|^{2}$;
      % \Else
      %   \State $x_{t-1}=x_{t-1}-A_t\times \nabla_{x_{t}}\left\| \acute{y}-k_{t}\otimes x_{t}^{0} \right\|^{2}$;
      % \EndIf
      \State $x_{t-1}=x_{t-1}'-A_t\times \nabla_{x_{t}}\left\| \acute{y}-k_{t}\otimes x_{t}^{0} \right\|^{2}$;
      \State $std_{t-1}=std_{t}-\sqrt{\bar{\alpha}_{t}}\times \nabla_{k_{t}}\left\| \acute{y}-k_{t}\otimes x_{t}^{0} \right\|^{2}$;
    \EndFor \\
  \Return $x_{0}$
  \end{algorithmic}
\end{algorithm}

%% file: sec/4_method.tex
\section{Method}
\label{sec:method}

% In this section, we introduce our method, DynFaceRestore, which use the Dynamic Blur-Level Mapping to remove degradation and restore the structure of unknown degraded LQ input based on degraded severity, the dynamic starting step lookup table to sample-wise determine starting timestep, and the dynamic guidance scaling adjuster to guide the sampling process in the region-wise view. 
% \phu{ In this section, we introduce our method, \textbf{DynFaceRestore}, which (1) employs a \textit{Dynamic Blur-Level Mapping} to restore structure and address degradation in unknown degraded low-quality (LQ) inputs, adapting to varying levels of degradation severity; (2) incorporates a \textit{Dynamic Starting Step Lookup Table} to determine the optimal starting timestep for each sample in the pre-trained diffusion model; and (3) utilizes a \textit{Dynamic Guidance Scaling Adjuster} to guide the sampling process with a region-wise focus.}

\CCH{
% In this section, we introduce \textbf{DynFaceRestore}, which includes three key components: (1) a \textit{Dynamic Blur-Level Mapping} (DBLM), which transforms blindly degraded low-quality (LQ) inputs into multiple Gaussian-blurred versions with blur levels adapted to the degradation level of the input; (2) a \textit{Dynamic Starting Step Lookup Table} (DSST), which identifies the optimal starting timestep to insert guidance into DM for each Gaussian-blurred image; and (3) a \textit{Dynamic Guidance Scaling Adjuster} (DGSA), which directs the diffusion sampling process in a region-wise manner. For an overview, we present the overall framework in Fig.\ref{fig:dynamicface} and outline the inference steps in Algorithm \ref{alg:inference}.   
}

%Fig.\ref{fig:dynamicface} illustrates the overall framework of the proposed method. These components are discussed in Sec.\ref{sec:Dynamic Blur-Level Mapping}, Sec.\ref{sec:Dynamic Starting Step}, and Sec.\ref{sec:Dynamic Guidance Scaling Estimator}. Finally, we conclude our method and introduce the inference steps in Algorithm \ref{alg:inference}.

\subsection{Dynamic Blur-Level Mapping}
\label{sec:Dynamic Blur-Level Mapping}

% \phu{As mentioned in Section \ref{sec:intro}, we model the output distribution of the Dynamic Blur-Level Mapping (DBLM) as a Gaussian blur distribution, leveraging the blur level, which is dynamically determined based on the degradation of the input $y$, to adapt the starting step. We train the DBLM using training pairs $[y, \tilde{x}]$ as follows:}

%\phu{The first step of our method involves mapping the degraded input images to Gaussian-blurred versions via the Dynamic Blur-Level Mapping (DBLM). Given the unknown degraded LQ input images $y$ and its HQ targets     images $x$, we optimize the DBLM by utilizing training pairs $[y, \tilde{x}]$ as follows:
%\Textbf{Dynamic Multi-Blur-Level Mapping (DMBM)}:
To leverage the reliable information (i.e., low-frequency components) of the degraded input and mitigate kernel mismatch issues during the diffusion process, we designed DBLM to first transform the degraded input $y$ into a Gaussian-blurred counterpart $\acute{y}$, effectively converting blind kernel to a Gaussian form. Next, $\acute{y}$ serves as a restored version of $y$, with maximally enhanced high-frequency details while preserving fidelity. Specifically, the DBLM function is defined as follows:
\begin{equation}
\begin{split}
\label{eq:DBLM_output}
\acute{y}=DBLM(y)=k^{std^*}_y\otimes RM(y),
\end{split}
\end{equation}
where $RM$ represents any state-of-the-art discriminative image restoration model, with SwinIR \cite{liang2021swinir} used in our implementation (details and other RMs' comparisons are defined in the Supplementary); $k^{std^*}_y$, parameterized by ${std^*}$, is the optimal Gaussian kernel related to the degradation level of $y$, where kernel standard ${std^*}$ is determined by
\begin{equation}
\label{eq:determine_std}
\begin{split}
%\tilde{x}&\equiv k^{std}\otimes x,  
%\quad &
% std^* &\equiv \underset{std\in[std_{min},std_{max}]}{arg min}(std), \\
% s.t. \quad \left\| & k^{std} \otimes RM(y)-k^{std}\otimes x \right. \left.\right\|_1<\xi.
std^* &\equiv \underset{std\in[std_{min},std_{max}]}{\arg\min}(std), \\
\text{s.t.} \quad & \left\| k^{std} \otimes RM(y)-k^{std}\otimes x \right\|_1 < \xi.
\end{split}
\end{equation}

In \cref{eq:determine_std}, $\xi$ denotes the chosen error tolerance, $x$ is the HQ ground truth, and $\otimes$ represents the convolution operation. Ideally, \cref{eq:determine_std} suggests that $\tilde{x}\equiv k^{std^*}_y\otimes x$ is the amount of information recoverable from $y$ by DBLM (as per \cref{eq:DBLM_output}) with fidelity. Moreover, since  $\acute{y}$ the output of DBLM is explicitly degraded by the Gaussian kernel $k^{std^*}_y$, utilizing $\acute{y}$ and $k^{std^*}_y$ as guidance within our diffusion-based restoration framework (\cref{sec:ours_guidance}) mitigates the impact of kernel mismatch issues caused by inaccurate kernel predictions. However, the estimation of ${std^*}$ merits further discussion during inference.
%: (a) and (b) dynamic multi-blur-Level mapping.

\subsubsection{Estimation of ${std^*}$}
\label{sec:se}
During inference, the ground truth $x$ is unavailable, necessitating an alternative approach for estimating ${std^*}$ instead of using \cref{eq:determine_std}. To achieve this, we design a learnable network (SE) to generate an estimate $\hat{{std^*}}$ from the input $y$. SE first produces an intermediate output $\hat{y}'$ that approximates $\tilde{x}$, and then, based on $\hat{y}'$ derives $\hat{{std^*}}$. In summary,  $[{\hat{std^*},\hat{y}'}]=SE(y)$. The training objective function for SE is defined as follows:
\begin{equation}
\begin{split}
\label{eq:SE_overall}
L_{SE}=\mathbb{D}(\hat{std^*},std^*)+\gamma_{std}\mathbb{D}(\hat{y}',\tilde{x})), 
\end{split}
\end{equation}
where $\mathbb{D}$ is the L1 distance and $\gamma_{std}$ is a weighting factor for balancing. Additionally, ${std^*}$ and ${\tilde{x}}$ are the training labels derived from \cref{sec:Dynamic Blur-Level Mapping}. 
\CCH{
Note that $\hat{y}'$, predicted by the SE, maps the unknown degraded input image ${y}$ to an intermediate blurred version, serving as an alternative to the Gaussian-blurred counterpart $\acute{y}$ from \cref{eq:DBLM_output}. However, $\hat{y}'$ does not perfectly conform to the assumed Gaussian form and is unsuitable as Gaussian-blurred guidance in the subsequent diffusion sampling process. Thus, in our final implementation, we use $\acute{y}$ instead of $\hat{y}'$, estimating $\acute{y}$ via \cref{eq:DBLM_output}, with $\hat{{std^*}}$ replaced by $\hat{{std^*}}$. Due to space constraints, a detailed analysis of the differences between $\acute{y}$ and $\hat{y}'$, along with the design details of the SE, is provided in the Supplementary.
}

\begin{figure}[t]
    \centering
    \includegraphics[width=\linewidth]{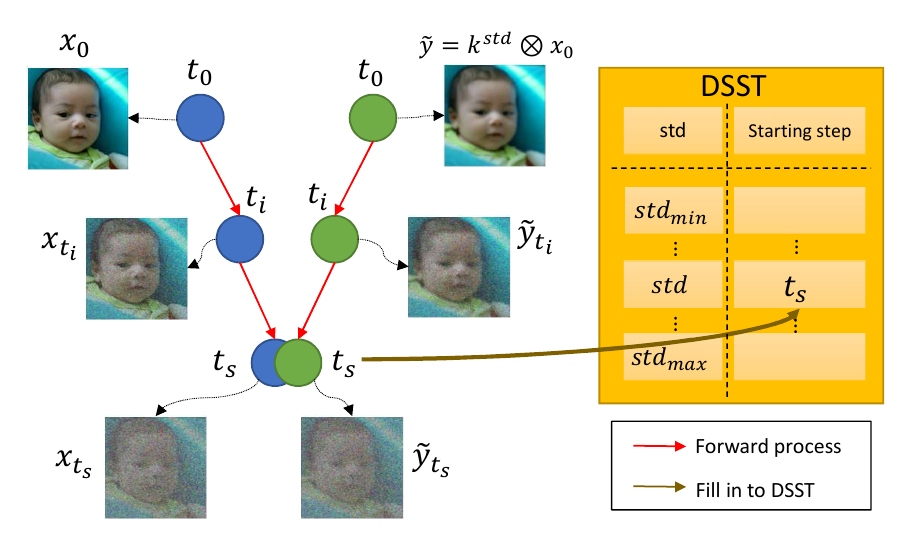}
    \caption{As $t$ increases, the diffused $x_{t}$ from HQ image $x_{0}$ and the diffused $\tilde{y}_{t}$ from the Gaussian-blurred image $\tilde{y}$ gradually approach and converge at $t_{s}$. We leverage this property to form the basis for establishing DSST.}
    \label{fig:dsst}
\end{figure}

\subsection{Dynamic Starting Step Lookup Table}
\label{sec:Dynamic Starting Step}
\CCH{
Based on the estimated blur level, $\hat{{std^*}}$, we develop a strategy to determine the starting timesteps for inserting guidance $\acute{y}$ into DM to prevent under- and over-diffusion. The idea is illustrated in \cref{fig:dsst}. Assume $x_{0}\equiv x$ represents an HQ image and $\tilde{y}_{0}^{std}= k^{std}\otimes x$ is its blurred counterpart. When both images undergo the same forward degradation process defined in a standard diffusion model,
%equation \ref{eq.ddpm_forward_xtx0},
they will statistically converge after a specific timestep $t$, meaning $x_{t}\cong \tilde{y}_{t}^{std}$. The timestep $t$ then serves as a suitable insertion point for a blurry observation $\tilde{y}_{0}^{std}$ with blur level $std$. To identify the optimal starting timestep $t_{std}$ for each standard deviation, we use the following equation:
\begin{equation}
\label{eq:SNR}
t_{std}=\underset{t}{argmin}\quad (log(\textbf{X}_{t})-log(\tilde{\textbf{Y}}_{t}^{std})\le tol),
\end{equation}
where $tol$ represents the maximum tolerance, $\textbf{X}_{t}$ and $\tilde{\textbf{Y}}_{t}^{std}$ denote the expected values of $x_{t}$ and $\tilde{y}_{t}^{std}$ in a training set.

Once we determine the corresponding starting timesteps for each standard deviation using \cref{eq:SNR}, we store these pairs in the Dynamic Starting Step Lookup Table (DSST). Given an estimated $\hat{{std^*}}$ via the SE network, the appropriate starting timestep can then be retrieved from the DSST.}

%\phu{Based on the blur level, $\hat{{std^*}}$, we construct a strategy to determine starting timesteps for inserting guidance into DM, avoiding under- and over-diffusion. Recall the forward process of DM, in equation \ref{eq.ddpm_forward_xtx0}, as $t$ increases, the added noise increasingly suppresses the high-frequency information in the original HQ image $x_{0}$. This characteristic implies that $\sqrt{\bar{\alpha_{t}}}x_{0}$ can closely resemble or even approximate $\sqrt{\bar{\alpha_{t}}}\tilde{y}^{std}$, where $\tilde{y}^{std}$ represents $x_{0}$ blurred by a Gaussian kernel $k^{std}$, as illustrated in Fig. \ref{fig:dsst}. Consequently, for each Gaussian blur kernel with a specified std, we aim to determine the timestep $t$ at which the noise from the forward process produces an $x_{t}$ that closely approximates the effect of applying similar noise to $\tilde{y}^{std}$. This relationship can be formulated as follows:}

% Once we identify the corresponding starting timesteps for each std using Equation \ref{eq:SNR}, we store these values in the Dynamic Starting Step Lookup Table (DSST). Since the output of DBLM is a Gaussian-blurred image based on the input degradation severity, it is possible to estimate the standard deviation of the DBLM output on a sample-wise basis using SE and query the corresponding starting timestep in DSST. This approach allows the avoidance of under-diffuse and over-diffuse while also shortening the sampling process.

\subsection{Dynamic Guidance Scaling Adjuster}
\label{sec:Dynamic Guidance Scaling Estimator}

\CCH{
Region-specific adjustment on guidance scale $s$ in \cref{eq.ddpm_backward_guidance} is crucial for diffusion sample refinement, as structured regions often require stronger diffusion (smaller $s$), while smoother regions need weaker diffusion (larger $s$). To address this, we propose the ``Dynamic Guidance Scaling Adjuster (DGSA)'', which adaptively adjusts $s$ on a region-wise basis throughout the DM sampling process. By dynamically modulating the guidance scale, DGSA minimizes over-smoothing by preserving structural integrity in areas such as facial contours while enhancing realistic details by reducing the scale in fine-detail regions like hair and wrinkles. With DGSA incorporated, the proposed guidance formula is then modified as follows:  
%This approach optimally utilizes the guidance while fully leveraging the DM’s facial generation capabilities, striking an ideal balance between fidelity and quality in the final output. 
}
\begin{equation}
\label{eq:DGSE}
x_{t-1}=
x_{t-1}'-A_{t}\times \nabla_{x_{t}}\left\| \acute{y}-k_{t}\otimes x_{t}^{0} \right\|^{2}, ~~ t\le t_{std}
\end{equation}
%\begin{equation}
%\label{eq:guidance}
%where\quad gd_{x_{t}}\equiv 
%\nabla_{x_{t}}\left\| \acute{y}-%k_{t}\otimes x_{t}^{0} \right\|^{2},
%\end{equation}
\begin{equation}
\label{eq:DGSE_input}
A_{t}\equiv 
DGSA(\acute{y},x_{t}^{0},t).
\end{equation}

% \begin{equation}
% \label{eq:DGSE}
% x_{t-1}=
% \left\{
% \begin{array}{lr}
% x_{t-1}-\sqrt{\bar{\alpha}_{t}}\times gd, & t>t_{fine} \\
% x_{t-1}-DGSA_{t}\times gd, & t\le t_{fine}
% \end{array}
% \right.
% \end{equation}

% \begin{equation}
% \label{eq:guidance}
% gd_{x_{t}}\equiv 
% \nabla_{x_{t}}\left\| \acute{y}-k_{t}\otimes x_{t}^{0} \right\|^{2},
% \end{equation}

% \begin{equation}
% \label{eq:DGSE_input}
% DGSA_{t}\equiv 
% DGSA(\acute{y},x_{t}^{0},t),
% \end{equation}

% \begin{equation}
% \label{eq:guidance}
% gd_{x_{t}}\equiv 
% \nabla_{x_{t}}\sum_{i=1}^{3}\lambda^{i}\left\| \acute{y}_{refine}^{i}-k_{t}^{i}\otimes x_{t}^{0} \right\|^{2},
% \end{equation}
% \begin{equation}
% \label{eq:DGSE_input}
% DGSA_{t}\equiv 
% DGSA(\acute{y}_{refine}^{1},x_{t}^{0},t),
% \end{equation}

%Here, $x_t^0$ is the HQ prediction from $x_t$
In \cref{eq:DGSE_input}, DGSA is a CNN with three convolution layers, producing an output in the range [0,1]. It takes as input the current measurement $\acute{y}$, the HQ prediction $x_t^0$ based on $x_t$, and $t$ to determine the local diffusion power needed at each timestep. The resulting output, $A_t$, is a region-wise guidance scaling map used to adjust the diffusion sampling.

%Nobally, $\acute{y}$ provides low-frequency information   

%we design a  the guidance is denoted as $gd$, and $x_t^0$ is the HQ prediction of $x_t$. The architecture of DGSA is two convolution layers followed by ELU, where ReLU-1 follows the last convolution layer to ensure the output is constrained to [0,1]. 

%The inputs to DGSA include $\acute{y}$, $x_t^0$, and $t$. DGSA utilizes $x_t^0$ to perform a region-wise search to identify areas where HQ information should be enhanced. At the same time, $\acute{y}$ provides low-frequency information to prevent high-quality details from being added to areas where it should preserve fidelity. Additionally, we embed the timestep $t$ as input to improve performance. 

% Consequently, the guidance scale map output by DGSA has the same dimensions as $x_t^0$, adapts the guidance scale region-wise, maintaining the scale in areas that require structural preservation, such as facial contours, while decreasing the scale in regions with fine details, such as hair and wrinkles.

\input{table/CelebA_Mix}

\begin{figure*}[t]
    \centering
    \includegraphics[width=\linewidth]{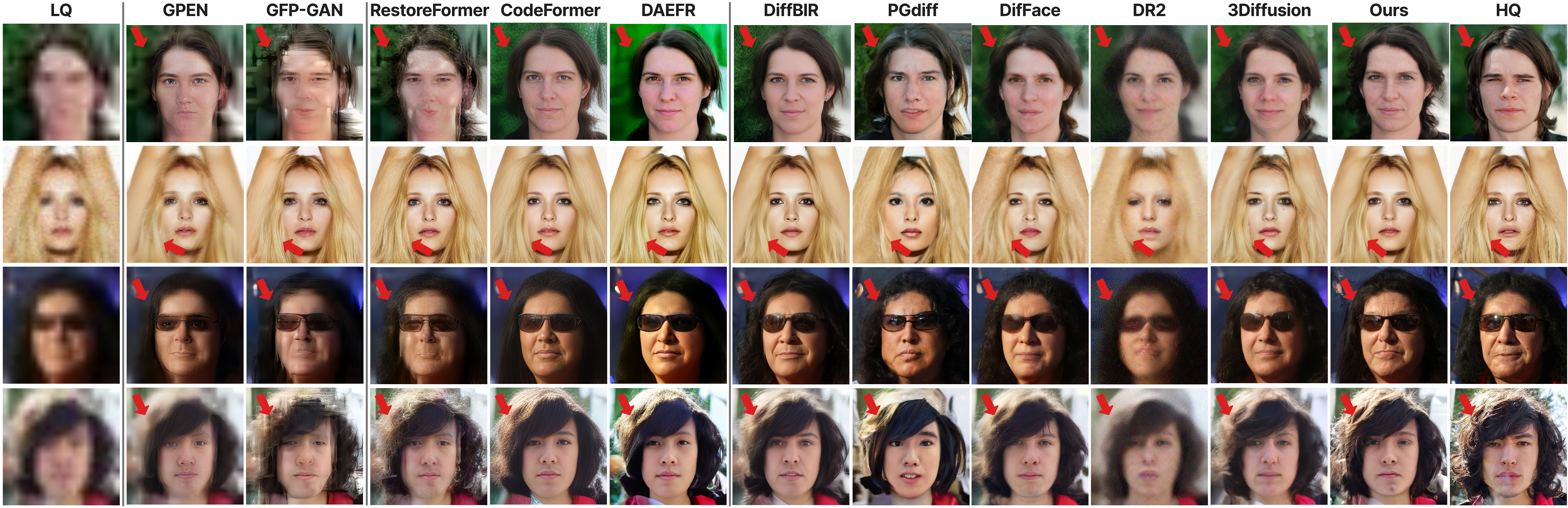}
    \caption{
    \CCH{Qualitative results on CelebA-Test. Our method achieves high-fidelity reconstruction with visually accurate details, particularly in the mouth, hair, and skin texture. Please zoom in for the best view.}
    % \caption{Qualitative results of CelebA-Test. GAN-based (GPEN, GFPGAN) and codebook-based methods (RestoreFormer, CodeFormer, DAEFR) suffer from fidelity issues, while PGDiff and DifFace exhibit deviations from the ground truth. In contrast, our DynFaceRestore delivers high-fidelity reconstructions.}
    }
    \label{fig:celeba-test}
    % \vspace{-5mm}
\end{figure*}

% To train the DGSA, we aim for the guidance-adjusted $x_{t}^{0}$ to resemble the ground truth. The training process follows the steps outlined in the Algorithm \ref{alg:DGSA}. After training, the DGSA learns to adapt the guidance scale region-wise appropriately, decreasing it in areas where fine details need to be generated, such as hair and skin, while maintaining or not significantly reducing the guidance scale in areas that need to preserve structure, such as facial contours.
The objective of training DGSA is to ensure that the HQ prediction $x^0_{t-1}$, based on $x_{t-1}$ (as described in \cref{eq.ddpm_backward}), closely approximates the ground truth $x_{0}$. Here, the adjusted sample $x_{t-1}$ is influenced by the DGSA network, as defined in \cref{eq:DGSE}. To achieve this, we randomly sample 
$t$ and use the following loss function to train DGSA:
%guidance-adjusted $x_{t-1}$ closely approximates the $x_{t-1}$ generated by adding $t-1$ steps of noise to the ground truth $x_{0}$ in any timestep. 
\begin{equation}
\begin{split}
    L_{DGSA}=  
    &\sum_{i}\gamma_{i}\mathbb{D}(SWT(x_{t-1}^{0})_{i},SWT(x_{0})_{i}])
    + \\ &DISTS(x_{t-1}^{0},x_{0}),
\end{split}
\label{eq:L_DGSA}
\end{equation}
where {$\gamma_{i}$} are the weighted factors of the four subbands (LL, LH, HL, HH) decomposed by Stationary Wavelet Transformation ($SWT$) \cite{jawerth1994swt,korkmaz2024wgsr}, and $DISTS$ \cite{ding2020dists,korkmaz2024wgsr} is the perceptual loss. We would described in detail the full training procedure of our method in the Supplementary .

% \phu{
% After training, the DGSA learns to adapt the guidance scale region-wise, maintaining the scale in areas that require structural preservation, such as facial contours, while decreasing the scale in regions with fine details, such as hair and eyes. This ensures that the inherent HQ facial priors in the DM can generate high-frequency details without excessive guidance towards the mean of $x_{0}$. This adaptive guidance scaling prevents the final sampled HQ facial image from becoming over-smoothed, allowing it to retain rich details while maintaining fidelity to the original input.}

\subsection{Multiple Mapping and Guidance}
\label{sec:Multiple Mapping}
Up to this point, we have employed only a single fidelity guidance, $\acute{y}$, and relied on the diffusion model (DM) to recover the missing high-frequency components (i.e., the difference between $x$ and $\acute{y}$) for restoration. However, while the DM effectively enhances image quality, it often sacrifices fidelity; conversely, the restoration model (RM) maintains fidelity but yields lower quality enhancement. To balance the strengths of DM and RM, we propose an optional add-on module that leverages multiple guidance sources. Specifically, once we obtain the estimated $\hat{{std^*}}$ from SE, we generate three Gaussian-blurred guidance images $\acute{y}^{i\in[1,2,3]}$ by varying $std$ values. Formally,
\begin{equation}
\begin{split}
\label{eq:DBLM_purturby}
\acute{y}^{i\in[1,2,3]}=k_y^{i\in[1,2,3]}\otimes RM(y),
\end{split}
\end{equation}
where $k_y^{i\in[1,2,3]}$ represents three Gaussian kernels with standard deviations $\hat{{std^*}}$, $\hat{{std^*}}-1$, and $\hat{{std^*}}-2$, respectively. These three guidance and corresponding kernels are then used in \cref{eq:DGSE} to update the diffusion sample. 

\input{table/Real}

Since $\acute{y}^{1}$ retains more trustable low-frequency structural information, while $\acute{y}^{2}$ and $\acute{y}^{3}$ provide more high-frequency details but with lower confidence, we prioritize adjustments derived from $\acute{y}^{1}$ by assigning it a higher weight, while assigning smaller weights to the guidance from $\acute{y}^{2}$ and $\acute{y}^{3}$. Additionally, the standard deviation of $k^{i\in[1,2,3]}$ are continuously adjusted throughout the diffusion process, as described in \cref{eq:backward_kernel_guidance}, which mitigates the impact of kernel mismatch and thereby enhances the fidelity of the final result. Nonetheless, one remaining issue is the selection of hyperparameters, including the number of guidance and the settings of $std$ values, which can be explored in future work.

%% file: table/CelebA_Mix.tex
\begin{table*}
\small
\centering
\caption{
Comparisons with SOTA methods on CelebA-Test. The best and second-best performances are highlighted in {\bf bold} and \underline{underline}. For fairness, all methods are re-evaluated on the same test sets. Computational performance is measured on an RTX 3090 GPU.
}
\label{tab:celeba_mix}
\scalebox{0.9}{
% \begin{tabular}{|c|c|c|c|c|c|c|c|lll}
\begin{tabular}{|c|c|c|c|c|c|c|c|c|c|c|}

    \hline
    % Type & Method & \makecell{Inference\\time (s)} & \makecell{MACs\\ (G)}  & \makecell{Params\\ (M)} &PSNR$\uparrow$ & SSIM$\uparrow$ & LPIPS$\downarrow$ & FID$\downarrow$ & IDA$\downarrow$ & LMD$\downarrow$  \\ 
    Type & Method & \makecell{Inference\\time (s)} & MACs (G)  & Params (M) &PSNR$\uparrow$ & SSIM$\uparrow$ & LPIPS$\downarrow$ & FID$\downarrow$ & IDA$\downarrow$ & LMD$\downarrow$  \\ 
    
    \hline
    \multirow{2}{*}{GAN-based}                    & GPEN &  0.368  & 562.998  & 129.203
&23.773& \underline{0.659}& 0.358& 30.250& 0.837& 6.377\\ 
    %\hline
                                        & GFP-GAN &  0.123  & 45.228  & 76.207
&22.841& 0.620& 0.355& 23.860& 0.822& 4.793\\
    \hline
    \multirow{3}{*}{Codebook-based}    & RestoreFormer &  0.152  & 343.051  & 72.680
&23.001& 0.592& 0.376& 22.874& 0.783& 4.464\\
    %\hline
                                    & CodeFormer &  0.062  & 297.206  &         94.113
&23.828& 0.637& \bf0.319& 18.076& 0.775& \underline{3.509}\\ 
    %\hline
                                     & DAEFR    &  0.288 & 455.605  & 152.151
& 21.640& 0.589& 0.345& 15.983& 0.870& 3.917\\
    \hline
    \multirow{5}{*}{DM-based}& DiffBIR    &  11.280  & 1525.137  &          1716.700
&24.127& 0.647& 0.357& 19.194& \underline{0.767}& 3.535\\
                                     & PGDiff   &  94.250  & 480.998  & 176.480
&21.824 & 0.612 & 0.369 & 20.928 & 0.944 & 4.868  \\
    %\hline
                                     & DifFace  &  6.120  & 272.678  &  175.430
&23.949& \underline{0.659} & 0.355 & \underline{15.032} & 0.867 & 3.781  \\
                                     & DR2      &  0.920  & 918.843  & 179.310
&21.023& 0.591& 0.461& 63.629& 1.183& 6.655\\
                                    & 3Diffusion&  11.759  & 308.417  &  180.510
&23.387& 0.651& 0.353& 15.446& 0.943& 3.781\\
    %\hline
    
       &        \cellcolor{Gray}  DynFaceRestore & \cellcolor{Gray} 91.820 & \cellcolor{Gray}323.048 & \cellcolor{Gray}176.770& \cellcolor{Gray}\bf24.349 & \cellcolor{Gray}\bf0.664 & \cellcolor{Gray}\underline{0.332} & \cellcolor{Gray}\bf14.780 & \cellcolor{Gray}\bf0.748& \cellcolor{Gray}   \bf3.419\\ 
       \hline
    % & \rowcolor{Gray} DynFaceRestore & \bf24.047 & \underline{0.653} & \underline{0.340} & \bf14.539 & 4.156 & 0.783 & \underline{3.615} \\
    % \hline
\end{tabular}
}
\end{table*}

% \begin{table*}
% \small
% \centering
% \caption{Comparisons to SOTA methods in CelebA-Test. The best and second performances are highlighted with {\bf bold} and \underline{underline}.}
% \label{tab:celeba_mix}
% \begin{tabular}{|c|c|c|c|c|c|c|c|c|}
%     \hline
%     Type & Method & PSNR$\uparrow$ & SSIM$\uparrow$ & LPIPS$\downarrow$ & FID$\downarrow$ & NIQE$\downarrow$ & IDA$\downarrow$ & LMD$\downarrow$ \\ 
%     \hline
%     \multirow{2}{*}{GAN} & GPEN & 23.444 & 0.638 & 0.374 & 25.662 & 4.688 & 0.789 & 5.919 \\ 
%     %\hline
%         & GFP-GAN & 22.971 & 0.623 & 0.352 & 23.401 & 4.132 & 0.801 & 4.693 \\
%     \hline
%     \multirow{3}{*}{Codebook}  & RestoreFormer & 23.140 & 0.596 & 0.374 & 22.660 & 4.279 & 0.763 & 4.361 \\
%     %\hline
%         & CodeFormer & 23.938 & 0.639 & \bf0.317 & 17.759 & 4.630 & \underline{0.759} & \bf3.443 \\ 
%     %\hline
%         & DAEFR & 21.715 & 0.591 & 0.343 & 15.827 & \underline{4.036} & 0.863 & 3.848 \\
%     \hline
%     \multirow{3}{*}{pretrained DM} & PGDiff & 21.824 & 0.612 & 0.369 & 20.928 & \bf3.955 & 0.944 & 4.868 \\
%     %\hline
%        & DifFace & \underline{23.949} & \underline{0.659} & 0.355 & \underline{15.032} & 4.451 & 0.867 & 3.781 \\
%     %\hline
%        & \rowcolor{Gray} DynFaceRestore & \bf24.177 & \bf{0.660} & \underline{0.334} & \bf14.356 & 4.239 & \bf0.740 & \underline{3.479} \\
%     \hline
%     % & \rowcolor{Gray} DynFaceRestore & \bf24.047 & \underline{0.653} & \underline{0.340} & \bf14.539 & 4.156 & 0.783 & \underline{3.615} \\
%     % \hline
% \end{tabular}
% \end{table*}

%% file: table/Real.tex
\begin{table}
\small
\centering
\caption{Comparisons to SOTA methods in real-world dataset (-Test). The best and second performances are highlighted with {\bf bold} and \underline{underline}.}
\label{tab:real}
\scalebox{0.9}{
\begin{tabular}{|c|c|c|c|c|}
    \hline
    \multirow{2}{*}{Type} & \multirow{2}{*}{Method} & LFW & WebPhoto & Wider \\ 
    \cline{3-5}
     & & FID$\downarrow$ & FID$\downarrow$ & FID$\downarrow$ \\
    \hline
     \multirow{2}{*}{GAN-based} & GPEN & 51.942 & 80.705 & 46.359 \\ 
        & GFP-GAN & 50.057 & 87.250 & 39.730 \\
    \hline
    \multirow{3}{*}{Codebook-based} & RestoreFormer & 48.412 & \underline{77.330} & 49.839 \\
        & CodeFormer & 52.350 & 83.193 & 38.798 \\ 
        & DAEFR & 47.532 & \textbf{75.453} & \underline{36.720} \\ 
    \hline
     \multirow{5}{*}{\makecell[c]{DM-based}}& DiffBIR& \underline{43.446}& 91.200& 36.721\\
 & PGDiff& 44.942& 84.163&40.380\\
    %\hline
       & DifFace & 48.688 & 87.811 & 37.510 \\
 & DR2& 70.979& 117.090&76.550 \\
 & 3Diffusion& 46.888& 83.335&37.130 \\
 & \cellcolor{Gray}DynFaceRestore &\cellcolor{Gray} \cellcolor{Gray}\textbf{42.515} & \cellcolor{Gray}95.317 & \cellcolor{Gray}\textbf{36.052} \\ 
    \hline
\end{tabular}
}
\vspace{-5mm}
\end{table}

% \begin{table*}
% \small
% \centering
% \caption{Comparisons to SOTA methods in real-world dataset. The best and second performances are highlighted with {\bf bold} and \underline{underline}.}
% \label{tab:real}
% \begin{tabular}{|c|c|c|c|c|c|c|c|}
%     \hline
%     \multirow{2}{*}{Type} & \multirow{2}{*}{Method} & \multicolumn{2}{c|}{LFW-Test} & \multicolumn{2}{c|}{WebPhoto-Test} & \multicolumn{2}{c|}{Wider-Test}  \\ 
%     \cline{3-8}
%      & & FID$\downarrow$ & NIQE$\downarrow$ & FID$\downarrow$ & NIQE$\downarrow$ & FID$\downarrow$ & NIQE$\downarrow$ \\
%     \hline
%      \multirow{2}{*}{GAN} & GPEN & 51.942 & 3.902 & 80.705 & 4.340 & 46.359 & 4.104 \\ 
%     %\hline
%         & GFP-GAN & 50.057 & 3.966 & 87.250 & 4.248 & 39.730 & 3.885 \\
%     \hline
%     \multirow{3}{*}{Codebook} & RestoreFormer & 48.412 & 4.168 & \underline{77.330} & 4.362 & 49.839 & 3.894 \\
%     %\hline
%         & CodeFormer & 52.350 & 4.482 & 83.193 & 4.708 & 38.798 & 4.164 \\ 
%         & DAEFR & 47.532 & \bf3.552 & \bf75.453 & \underline{4.041} & \underline{36.720} & \bf3.655 \\ 
%     \hline
%      \multirow{3}{*}{pretrained DM} & PGDiff & \underline{44.942} & \underline{3.774} & 84.163 & \bf3.928 & 40.380 & \underline{3.827} \\
%     %\hline
%        & DifFace & 48.688 & 4.219 & 87.811 & 4.373 & 37.510 & 4.391 \\
%     % \cline{2-8}
%     %    & \rowcolor{Gray} DynFaceRestore & \bf43.305 & 3.932 & 87.846 & 5.003 & \bf36.130 & 3.928 \\
%     % \hline
%     & \rowcolor{Gray} DynFaceRestore & \bf44.372 & 4.745 & 90.058 & 5.361 & \bf36.549 & 4.042 \\
%     \hline
% \end{tabular}
% \end{table*}

%% file: sec/5_experiments.tex
\CCH{
\section{Experiments}
\label{sec:experiment}

\begin{figure*}[t]
    \centering
    \includegraphics[width=1.0\linewidth]{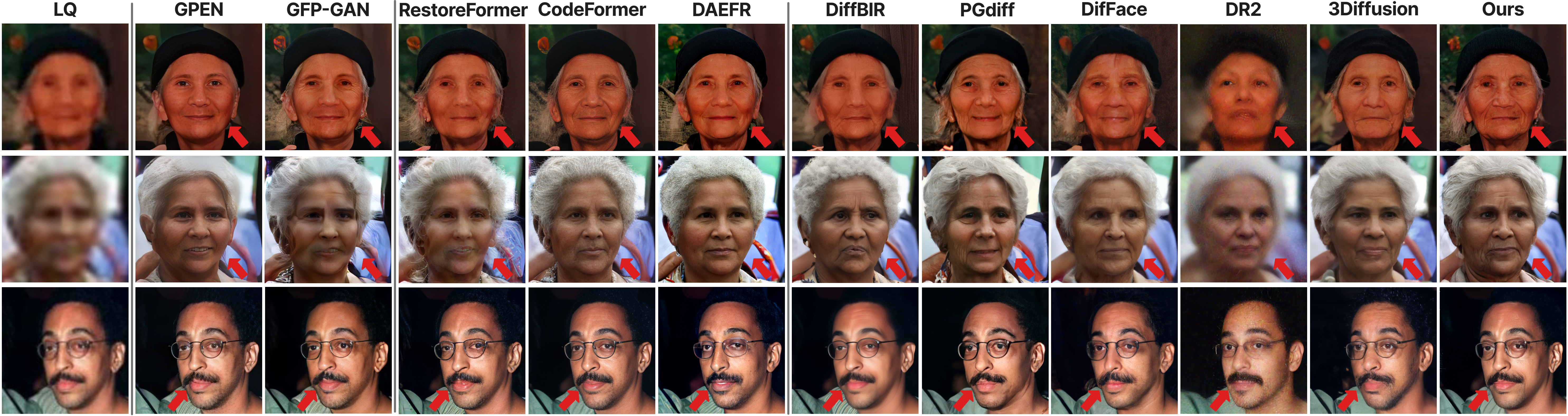}
    \caption{Qualitative results from three real-world datasets demonstrate that our restoration method produces more natural features (e.g., eyes) and realistic details (e.g., hair) compared to other approaches, with improved fidelity. Please zoom in for the best view.}
    \label{fig:real-world}
\end{figure*}

\subsection{Datasets and Metrics}
\noindent\textbf{Training dataset.} Our framework is trained on the FFHQ dataset \cite{karras2019ffhq}, which provides HQ facial images as the baseline. To simulate real-world degradation and generate LQ counterparts, we apply the following degradation pipeline:
\begin{equation}
\label{eq:degradation_pipeline}
y=\left\{ JPEG_{\delta}[(x\otimes k_{\sigma})\downarrow _{C}+n_{\zeta}] \right\}\uparrow _{C}.
\end{equation}
Here, $\sigma$ represents the standard deviation of the blur kernel, uniformly sampled from [0.1, 15]. The scale factor $C$, used for both downsampling and upsampling, is uniformly sampled from [0.8, 32]. The noise level $\zeta$ is also uniformly sampled from [0, 20], and $\delta$ denotes the JPEG compression quality factor, uniformly sampled from [30, 100]. We use the same hyperparameters settings as \cite{miao2024waveface}.

% where $\sigma$ is the standard deviation of the blur kernel uniform sampling from [0.1, 15], $s$ is the downsampling and upsampling scale uniform sampling from [0.8, 32], $\zeta$ is the noise level uniform sampling from [0, 20], and $\delta$ is the JPEG compression quality factor uniform sampling from [30, 95].
% where $\sigma$ is the standard deviation of the blur kernel, $s$ is the downsampling and upsampling scale, $\zeta$ is the noise level, and $\delta$ is the JPEG compression quality factor.
% The training procedure details are provided in the supplementary material.

\noindent\textbf{Testing datasets} We evaluate our method on one synthetic dataset (CelebA) and three real-world datasets (LFW, WebPhoto, and Wider). The CelebA-Test dataset comprises 3,000 LQ facial images synthesized from the CelebA-HQ \cite{karras2017celebahq} validation set using the degradation pipeline outlined in \cref{eq:degradation_pipeline}. For real-world testing, LFW-Test \cite{huang2008lfw}, consists of 1,711 slightly degraded images, each representing a unique individual from the LFW dataset; WebPhoto-Test \cite{kumar2022gfpgan} comprises 407 internet-sourced images, including heavily degraded old photos; and Wider-Test \cite{zhou2022codeformer} features 970 severely degraded images from the Wider dataset \cite{yang2016wider}. These datasets, encompassing varying levels of degradation, enable a comprehensive evaluation of our method’s robustness and effectiveness in real-world scenarios.

\noindent\textbf{Evaluation metrics} PSNR and SSIM are used for the synthetic dataset to assess fidelity, while LPIPS \cite{zhang2018lpips} measures perceptual quality. IDA evaluates identity consistency based on embeddings from ArcFace \cite{deng2019arcface}, and LMD \cite{tsai2024dualassociatedencoderface} assesses facial alignment and expression accuracy. Finally, FID \cite{heusel2017fid} is employed to evaluate perceptual quality. For the synthetic testing case, the CelebA-HQ dataset is used as the reference set; for real-world testing, the FFHQ dataset forms the reference set.

%using the CelebA-Test as the target set and FFHQ datasets as reference sets, respectively.
%  CelebA-HQ

To ensure a fair comparison, we utilized the same pre-trained diffusion model as DifFace \cite{yue2024difface} and PGDiff \cite{yang2024pgdiff}. Due to space limitations, comprehensive implementation details, hyperparameter configurations, training procedures, and additional results are provided in the Supplementary.
}

% For a fair comparison, we use the same pre-trained diffusion model as DifFace \cite{yue2024difface} and PGDiff \cite{yang2024pgdiff}. 
% We assign $\lambda^{[1,2,3]}$ as 0.7, 0.2, and 0.1 for the weights of multiple guidance. 
% The implementation details, hyperparameter settings, and the training procedure are provided in detail in the Supplementary.

% \subsection{Implementation Details}
% For fair comparison, the pre-trained diffusion model we used is borrowed from DifFace \cite{yue2024difface}, which was trained on the FFHQ dataset. The architecture of DBLM is also as same as diffused estomator used in DifFace, namely SwinIR \cite{liang2021swinir}. The std estimator was trained for 100,000 iterations, with $std_{max}$ set to 15.0 and $std_{min}$ set to 0.1. We used Equation \ref{eq:degradation_pipeline} to synthesize the low-quality (LQ) face images and applied Equation \ref{eq:blur_gt} to find the corresponding blurry ground truth for each LQ face, forming training pairs to train the DBLM. The DBLM was trained for 150,000 iterations, with the learning rate set to 2e-4, halving every 50,000 iterations. During this training, the std estimator was frozen, and $\lambda_{std}$ was set to 1e-2. 

% The tolerance $tol$ was set to 1e-3 for creating the Dynamic Starting Step Look-up Table (DSST) for following experiments. The training of DGSA takes 200,000 iterations, with a learning rate of 1e-4. In the inference, we empirically set the $t_{fine}$ to 400. 

\CCH{
\subsection{Comparisons to SOTA methods}
We conduct comprehensive quantitative and qualitative comparisons of our proposed DynFaceRestore with the following state-of-the-art BFR methods: $GAN-based$ (GPEN \cite{yang2021gpen}, GFPGAN \cite{kumar2022gfpgan}), $Codebook-based$ (RestoreFormer \cite{wang2022restoreformer}, CodeFormer \cite{zhou2022codeformer}, DAEFR \cite{tsai2024dualassociatedencoderface}) and $Diffusion Model (DM)-based$ (DiffBIR \cite{diffbir}, PGDiff \cite{yang2024pgdiff}, DifFace \cite{yue2024difface}, DR2 \cite{dr2} and 3Difffusion \cite{3Diffusion}).
}

\noindent\textbf{Comparison on Synthetic Dataset.} \cref{tab:celeba_mix} presents the quantitative results on the CelebA-Test dataset, demonstrating our method outperforms others in PSNR, SSIM, FID, IDA, and LMD metrics, while ranking second in LPIPS. These results highlight the method’s ability to balance fidelity and perceptual quality well. 
% These outcomes underscore the effectiveness of our approach in balancing fidelity and perceptual quality.

\CCH{
Qualitative results in \cref{fig:celeba-test} further highlight the advantages of our method. GAN-based approaches often introduce artifacts, while codebook-based methods generate high-quality details but still suffer from fidelity issues. Similarly, diffusion-based methods exhibit deviations from the ground truth. In contrast, our method consistently achieves high-fidelity reconstructions, preserving critical features such as the mouth, hair, and skin texture while maintaining image quality across varying degradation levels. This demonstrates its ability to produce visually accurate results.

}

\input{table/ablation}

\begin{figure}[t]
    \centering
    \includegraphics[width=\linewidth]{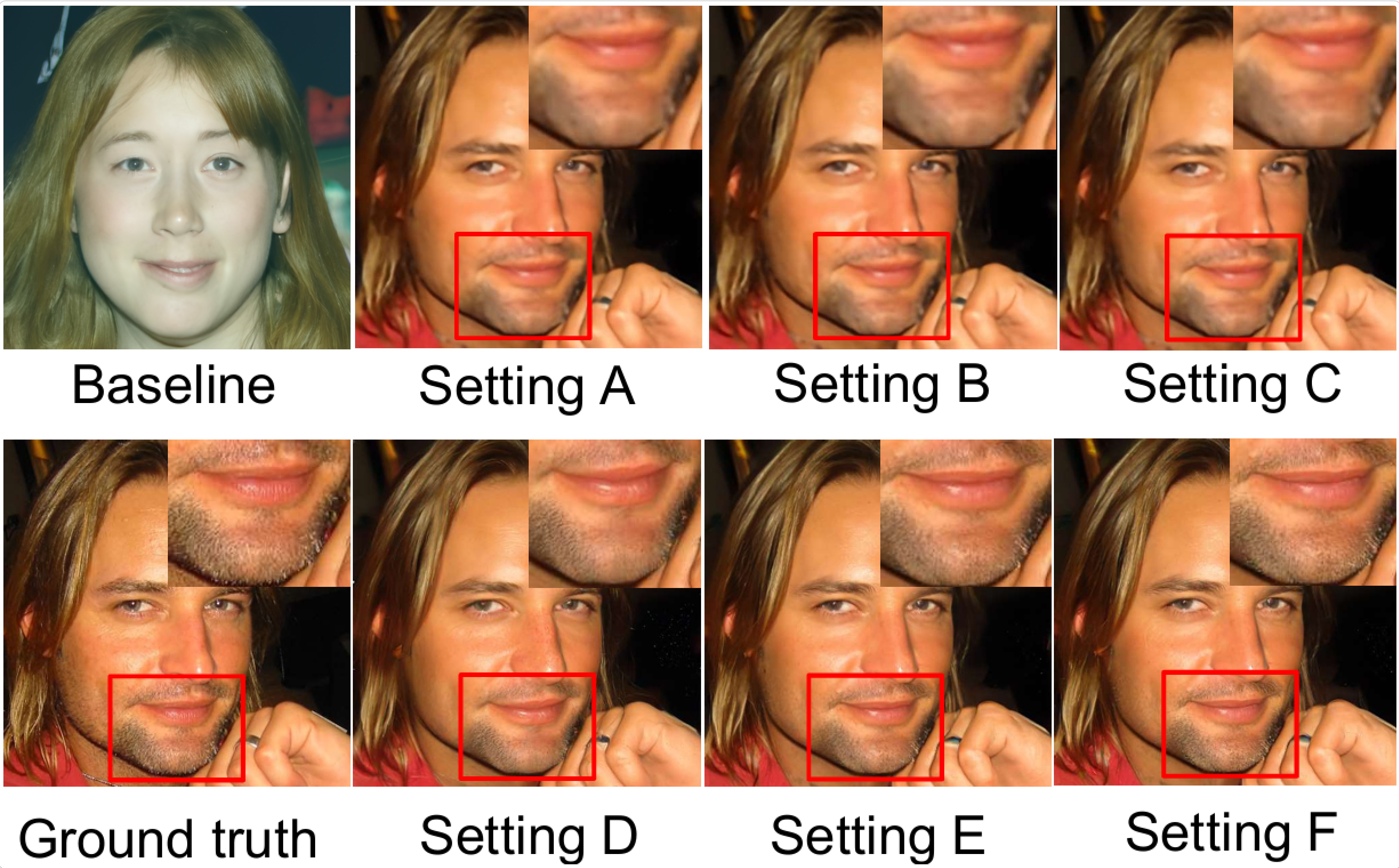}
    \caption{ Ablation results for each component in DynFaceRestore are shown, using the experimental settings in \cref{tab:ablation} for visual comparison. With the addition of DBLM, DSST, DGSA, and multiple guidance, DynFaceRestore well balances fidelity and quality.   
    }
    \label{fig:ablation}
    \vspace{-5mm}
\end{figure}

\noindent\textbf{Comparison on Real-world Datasets.} In \cref{tab:real}, we provide quantitative results across three real-world datasets, showing that our method surpasses competing approaches by achieving the best FID scores on the LFW-Test and the Wider-Test. For the WebPhoto-Test, it is worth noting that the FID score may not accurately reflect the discrepancy from the actual data distribution, given the dataset’s smaller sample size of 407 images, as also noted in prior works such as DifFace \cite{yue2024difface} and WaveFace \cite{miao2024waveface}.

\CCH{
\cref{fig:real-world} showcases qualitative results on these real-world datasets. GAN-based methods often produce artifacts that compromise the quality crucial for BFR. While Codebook-based and DM-based methods demonstrate improvements in realism, our approach achieves higher perceptual quality across varying degradation levels. These observations show the ability to produce high-quality details of  DynFaceRestore.
% This is mainly due to the tailored starting step selection and the implementation of region-specific guidance scale adjustments, collectively ensuring superior restoration quality.
}

% \begin{figure}[ht]
%     \centering
%     \includegraphics[width=\linewidth]{figure/DGSA.pdf}
%     \caption{Visualization of DGSA in various timestep through the sampling process. The guidance scale is smaller when leaning toward blue and larger when leaning toward red. As $t$ gradually decreases, areas requiring DM's HQ prior to adding details, such as hair, exhibit a lower guidance scale. In contrast, regions on the face maintain a higher guidance scale to preserve fidelity.}
%     \label{fig:dgsa}
% \end{figure}

\subsection{Ablation Study}
We evaluate various model aspects depending on different settings ($A, B, C, D, E, F$) as shown in \cref{tab:ablation} and \cref{fig:ablation}, along with a $Baseline$ comparison laveraged diffusion model \cite{dhariwal2021diffusion} trained on the FFHQ \cite{karras2019ffhq}.

% The qualitative results are further shown in \cref{fig:ablation}. 

% \noindent\textbf{Effect of DBLM.} Compared to the $Baseline$, setting $A$, achieves better performance in PSNR, preserving fidelity by relaxing BFR task to Gaussian deblurring and effectively applying guidance. As shown in \cref{fig:ablation}, the mouth and beard in setting $A$ are closer to GT.
\CCH{
\noindent\textbf{Effect of DBLM.} Compared to the $Baseline$, setting \textbf{A} achieves remarkable improvements across all metrics by reframing the BFR task as Gaussian deblurring and incorporating DM guidance. As shown in \cref{fig:ablation}, the mouth and beard regions in setting \textbf{A} are closer to Ground Truth.
% However, as mentioned in \cref{sec:Multiple Mapping}, the absence of multiple guidance shifts reliance toward the DM for enhancing details, resulting in lower PSNR but improved FID.
% \input{table/DFRM}

\noindent\textbf{Effect of Multiple Guidance.} Setting $B$ and $F$ further enhance fidelity over $A$ and $E$, respectively, improving both PSNR and IDA metrics by adaptively incorporating multiple guidance sources. Specifically, three Gaussian-blurred images are applied to guide the diffusion sampling process in this experiment.

% As also shown in \cref{tab:DFRM}, increasing the number of guidance enhances fidelity, as demonstrated by the improvements in both PSNR and IDA metrics.
}

\noindent\textbf{Effect of DSST.} Compared to setting $B$, integrating the DSST table in setting $C$ improves both PSNR and FID, while also shortening the sampling process, as reflected in the reduced range of sampling steps. This improvement is achieved by identifying the optimal starting step, effectively avoiding under- or over-diffusion.

\noindent\textbf{Effect of DGSA.} Setting $D,E,F$ achieves a notable improvement in the perceptual quality (FID score) compared to settings $A,B,C$, as illustrated in \cref{fig:ablation} and \cref{tab:ablation}. 
This demonstrates the effectiveness of the region-wise guidance scale. Visual analysis of the scaling map is provided in the Supplementary to explain the DGSA's behavior.

% Along with a reduction in DM steps, 
Our approach provides dynamic control, ranging from enhancing fidelity restoration (setting $C$) to effectively improving image quality, achieving the best FID score (setting $E$). This highlights the flexibility of our model,  allowing us to balance perceptual quality and fidelity by managing guidance resources (setting $F$).

% We further illustrate the output of DGSA in Fig.\ref{fig:dgsa} to demonstrate its effectiveness. As the timestep decreases, DGSA adjusts the guidance scale regionally. For regions like the hair and wrinkles, where finer details are required, the guidance scale gradually decreases, allowing these areas to leverage more of the pre-trained DM’s HQ image prior during the sampling process to generate realistic facial details. Conversely, the guidance scale remains high for structural regions to preserve the structure, ensuring that the final output stays close to the ground truth. This balance results in outputs that maintain high fidelity and rich facial details.

%% file: table/ablation.tex
\begin{table*}
\small
\centering
\captionsetup[table]{skip=0pt}
\caption{Ablation study of our framework on CelebA-Test. The best performance is highlighted with {\bf bold}.}
\label{tab:ablation}
\scalebox{0.9}{
\begin{tabular}{|l|c|c|c|c|c|l|c|l|c|} \hline 
% \begin{tabular}{|l|c|c|c|c|c|l|c|c|c|c|c|l} \hline 
    
    Setting & $DBLM$ & $Multiple$ $Guidance$ & $DSST$ & $DGSA$ & PSNR$\uparrow$ &  SSIM$\uparrow$&FID$\downarrow$  &IDA$\downarrow$& Range of sampling steps \\ \hline  
    
%     Baseline (DifFace) & & & & & 23.949 &  0.659
% &15.032  &0.869
% & 400 \\
 Baseline (DM)& & & & & 11.129& 0.382& 55.777& 1.461&1000\\   
    % \hline
    A & \checkmark & & & & 24.992 &  0.690&18.303  &0.725
& 1000 \\ 
    %\hline
    B & \checkmark & \checkmark & & & 25.094 &  0.694
&19.823  &0.725
& 1000 \\ 
    %\hline
    C & \checkmark & \checkmark & \checkmark &  & \bf25.107 &  \bf0.695&19.786  &\bf0.724& [690, 925] \\ 
    % \hline
    D & \checkmark & \checkmark  & & \checkmark &  24.340 &  0.664
&15.007  &0.750
& 1000 \\ 
    %\hline
    E & \checkmark &  & \checkmark & \checkmark & 24.327 &  0.666
&\bf14.689  &0.755
& [690, 925] \\ 
    %\hline
    \cellcolor{Gray} F & \cellcolor{Gray} \checkmark & \cellcolor{Gray} \checkmark & \cellcolor{Gray} \checkmark & \cellcolor{Gray} \checkmark & \cellcolor{Gray} 24.349 &  \cellcolor{Gray} 0.664& \cellcolor{Gray} 14.780  & \cellcolor{Gray} 0.748& \cellcolor{Gray} [690, 925] \\
     %\hline
    % F (Ours) & \checkmark &  & \checkmark & \checkmark & 24.327 & \bf14.689 & [690, 925] \\ \hline 
    
\end{tabular}
}
\end{table*}

% \begin{table*}
% \small
% \centering
% \caption{Ablation study of our framework on CelebA-Test. The best performance is highlighted with {\bf bold}.}
% \label{tab:ablation}
% \begin{tabular}{|l|c|c|c|c|c|c|}
%     \hline
%     Setting & $DFRM$ & $DSST$ & $DGSA$ & PSNR$\uparrow$ & NIQE$\downarrow$ & Avg. \# of NFE \\ 
%     \hline
%     Baseline (DifFace) & & & & 23.949 & 4.451 & $0.4T$  \\ 
%     %\hline
%     A & \checkmark & & & 24.760 & 4.684 & $T$  \\ 
%     %\hline
%     B & \checkmark & \checkmark & & \bf24.780 & 4.647 & $0.85T$  \\
%     % \hline
%     C & \checkmark & & \checkmark & 24.044 & 4.210 & $T$ \\
%     %\hline
%     D (DynFaceRestore) & \checkmark & \checkmark & \checkmark & 24.047 & \bf4.156 & $0.85T$  \\ 
%     \hline
% \end{tabular}
% \end{table*}

%% file: sec/6_conclusion.tex
\vspace{-2mm}
\section{Conclusion}
\vspace{-2mm}
% We propose DynFaceRestore, a novel approach to address Blind Face Restoration (BFR). We simplify the task through Dynamic Blur-Level Mapping by mapping unknown degraded input to a corresponding Gaussian blur version, transforming BFR into the Gaussian deblurring problem. DynFaceRestore achieves an optimal balance between fidelity and perceptual quality by selecting tailored starting steps and applying region-specific guidance scaling, which preserves identity while enriching details. Our method also incorporates multiple guidance mechanisms to further bolster fidelity. Experiment results demonstrate that DynFaceRestore exhibits strong generalization and superior performance.

% We propose DynFaceRestore, a novel approach to address Blind Face Restoration (BFR). We simplify the task through Dynamic Blur-Level Mapping by mapping unknown degraded input to a corresponding Gaussian blur version, transforming BFR into the Gaussian deblurring problem. DynFaceRestore achieves impressive overall performance by adaptively selecting tailored starting steps, adjusting the guidance scale with region-specific scaling, and providing flexible control to achieve an optimal balance between fidelity and perceptual quality through multiple guidance sources. Experiment results demonstrate that DynFaceRestore exhibits strong generalization and superior performance.

We propose DynFaceRestore for Blind Face Restoration (BFR). By utilizing Dynamic Blur-Level Mapping, we simplify the task by mapping unknown degraded input to a corresponding Gaussian blurred image, effectively reframing BFR as a Gaussian deblurring problem. Combined with optimal starting steps and region-wise scale adjustments, our method delivers superior performance. Experimental results demonstrate that DynFaceRestore outperforms existing methods while striking a balance between fidelity and perceptual quality. Additional experiments, comparisons, and analyses are provided in the Supplementary.
%, enabled by a flexible control mechanism leveraging multiple guidance sources.

%ur framework also ensures an optimal balance between fidelity and perceptual quality through a flexible control mechanism of multiple guidance sources.

% We propose DynFaceRestore, a novel approach to address Blind Face Restoration (BFR). We simplify the task through Dynamic Blur-Level Mapping by mapping unknown degraded input to a corresponding Gaussian blur version, transforming BFR into the Gaussian deblurring problem. By combining this with optimally identified starting steps and region-specific scale adjustments, DynFaceRestore achieves impressive performance.
% Additionally, our framework provides flexible control, allowing for an optimal balance between fidelity and perceptual quality by leveraging multiple guidance sources. Experiment results demonstrate that DynFaceRestore exhibits strong generalization and superior performance.

%% file: sec/X_suppl.tex
% \clearpage
\maketitlesupplementary

\section{Implementation Details}
\label{sec:imple_details}

\subsection{Dynamic Blur-Level Mapping}

\noindent\textbf{Architecture:}
With the objective of transforming the degraded input $y$ into a Gaussian-blurred counterpart $\acute{y}$, the proposed Dynamic Blur-Level Mapping (DBLM) is composed of two primary modules: the Standard Estimation ($SE$) module and the Restoration Model ($RM$). 
Specifically, the $SE$ predicts the Gaussian blur standard deviation $\hat{std^*}$ corresponding to $y$, while the $RM$ refines $y$ into the high-quality (HQ) distribution. The final Gaussian-blurred output, as described in Eq. (8) of the main paper, is rewritten as follows:
\begin{equation}
\label{eq:dblm_outputoutput}
\begin{split}
\acute{y}=k^{\hat{std^*}}\otimes RM(y),
\end{split}
\end{equation}
where $RM$ represents any SOTA pre-trained restoration model capable of mapping degraded inputs directly to high-quality (HQ) outputs. In our experiments, we adopt SwinIR \cite{liang2021swinir} architecture, which has been well-trained in DifFace \cite{yue2024difface}. $k^{std^*}$ denotes as Gaussian blur kernel defined by $\hat{std^*}$ estimated by $SE$ module.  
% The DBLM output is generated by convolving the Gaussian blur kernel, defined by the blur level predicted by the SE, with the RM’s output, ensuring consistency with the target Gaussian blur distribution.

\begin{figure}[t]
\centering    \includegraphics[width=\linewidth]{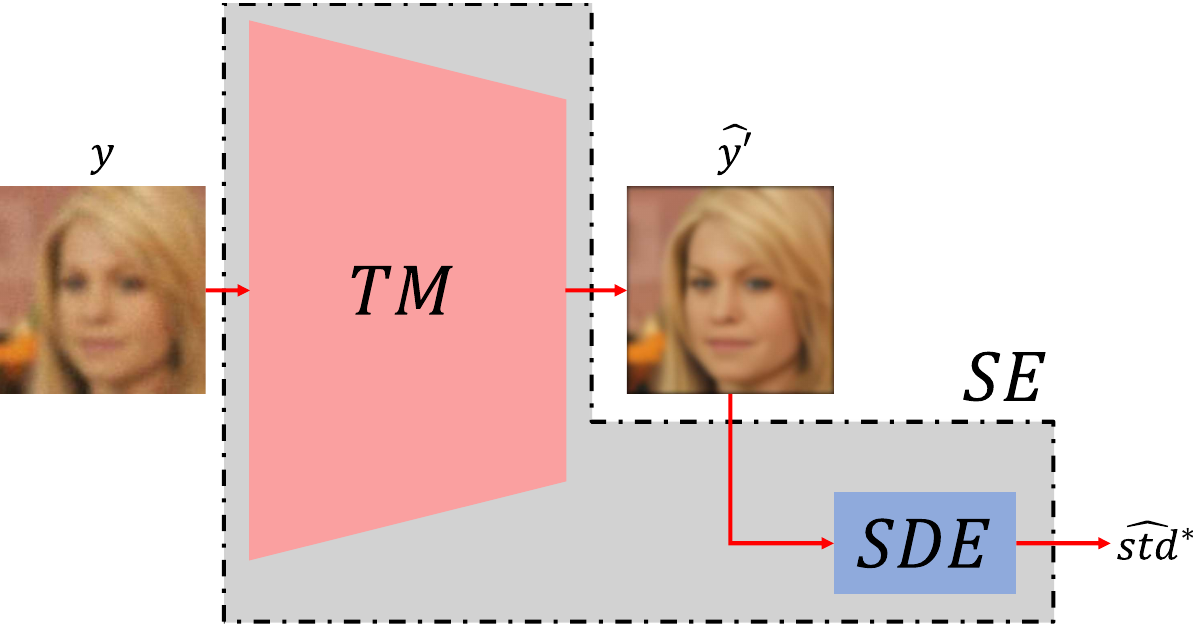}
    \caption{The SE consists of $TM$ and a $SDE$, predicting the Gaussian blur level corresponding to the degraded input based on intermediate Gaussian blur image $\hat{y}'$.}
    \label{fig:SE}
\end{figure}

% Để ước tính chính xác ${\hat{std^*}$ during the inferecne, we thiết kế 
As shown in \cref{fig:SE}, the $SE$ comprises a Transfer Model ($TM$) and a Standard Deviation Estimator ($SDE$). The $TM$, built on the SwinIR architecture \cite{liang2021swinir}, converts the degraded input $y$ into an intermediate Gaussian-blurred image $\hat{y}'$. Subsequently, the $SDE$
%, inspired by the degradation encoder of DASR \cite{wang2021dasr}, 
processes this intermediate image to estimate the corresponding Gaussian blur level $\hat{std^*}$.
The detailed architecture of the $SDE$ is provided in \cref{tab:arch_se}. 
% The $TM$ adopts an architecture based on SwinIR \cite{liang2021swinir}, while the $SDE$ incorporates a degradation estimation module inspired by DASR \cite{wang2021dasr}. Detailed architectural descriptions are available in \cref{tab:arch_se}.

% The RM is responsible for transforming the degraded input into an HQ distribution. Notably, the RM architecture is flexible, allowing any SOTA restoration model capable of mapping degraded inputs directly to HQ outputs. In our experiments, the RM architecture is based on SwinIR \cite{liang2021swinir}.

\noindent\textbf{Data Preparation for Training $SE$:} 
To train the $SE$ module and implement the concept of DBLM, we first prepare the training labels. Specifically, for synthesized low-quality (LQ) facial images, we collect pairs of ground truth labels: (1) kernel standard deviations ${std^*}$ to supervise the $SDE$ and (2) Gaussian-blurred images $\tilde{x}$, generated by degrading the high-quality (HQ) image $x$ using $k^{std^*}$, to supervise the $TM$. The synthesized low-quality (LQ) facial images $y$ are created using the degradation pipeline described below:

\begin{equation}
\label{eq:degradation_pipeline}
y=\left\{ JPEG_{\delta}[(x\otimes k_{\sigma})\downarrow _{C}+n_{\zeta}] \right\}\uparrow _{C}.
\end{equation}
Here, we use the same hyperparameter settings as \cite{miao2024waveface} for synthesizing low-quality (LQ) facial images. Additionally, the kernel standard deviation label  ${std^*}$ is determined by using \cref{eq:determine_std}, and the corresponding Gaussian-blurred ground truth is generated using $\tilde{x} \equiv k^{\text{std}^*}_y \otimes x$.
% $\tilde{x}&\equiv k^{std^*}_y\otimes x$. 

%to determine the  for each LQ input, forming training pairs to train the $TM$ module within the SE:
% \begin{equation}
% \label{eq:determine_std}
% \begin{split}
% std^* &\equiv \underset{std\in[std_{min},std_{max}]}{arg min}(std), \\
% s.t. \quad \left\| & k^{std} \otimes RM(y)-k^{std}\otimes x \right. \left.\right\|^{1}<\xi.
% \end{split}
% \end{equation}

\begin{equation}
\label{eq:determine_std}
\begin{split}
std^* &\equiv \underset{std\in[std_{min},std_{max}]}{\arg\min}(std), \\
\text{s.t.} \quad & \left\| k^{std} \otimes RM(y)-k^{std}\otimes x \right\|^{1} < \xi.
\end{split}
\end{equation}

Here, $\xi$ represents the error tolerance, $x$ denotes the high-quality (HQ) ground truth, and $\otimes$ signifies the convolution operation. The search space for the standard deviation range, $[std_{min}, std_{max}]$, is defined between 0.1 and 15.0 with an interval of 0.1. To solve the optimization problem in \cref{eq:determine_std}, we employ a brute-force method.

%However, configuring the hyperparameter $\xi$ presents a significant challenge. When $\xi$ is set too large, $k^{std} \otimes RM(y)$ deviates substantially from $k^{std} \otimes x$. Conversely, setting $\xi$ too small results in an overestimated $std^*$, leading to the loss of valuable information from the original input $y$. We extend the concept in \cref{eq:determine_std} by introducing an alternative constraint to address this limitation. This approach leverages that $RM(y)$ and $x$ share the same low-frequency information from $k^{std^*} \otimes x$. We modify the constraint to determine $std^*$ by maximizing the information from the input $y$ retained in $k^{std^*} \otimes x$:
% \begin{equation}
% \label{eq:blur_gt}
% \begin{split}
% std^* &\equiv \underset{std\in[std_{min},std_{max}]}{\mathrm{arg\,min}}(\left\| y-k^{std}\otimes x \right. \left.\right\|^{1}).
% \end{split}
% \end{equation}
%This alternative constraint allows for the effective utilization of reliable information from the input while avoiding the hallucination of unreliable details, thereby enhancing the fidelity of the restored image.

\noindent\textbf{Training procedure of $SE$:} 
The training procedure of SE begins with $SDE$ within the SE. Here, Gaussian-blurred images are generated by \cref{eq:y_dot} and used to train the $SDE$ by $L_{SDE}$ defined in \cref{eq:loss_SDE}.

% \begin{equation}
% % \begin{split}
% \label{eq:y_dot}
% \ddot{y}&=k^{\ddot{std}}\otimes x.
% % \end{split}
% \end{equation}

\begin{equation}
% \begin{split}
\label{eq:y_dot}
\ddot{y}=k^{\ddot{std}}\otimes x.
% \end{split}
\end{equation}

$L_{SDE}$ denotes the loss function that aligns the $SDE$’s predictions with the actual blur levels $\ddot{std}$. After completing this training phase, the $SDE$ reliably estimates the Gaussian kernel for any Gaussian-blurred image, forming a robust foundation for effective blur-level prediction in the subsequent stages of the DBLM framework.

% \begin{equation}
% % \begin{split}
% \label{eq:loss_SDE}
% L_{SDE}&=\mathbb{D}(SDE(\ddot{y}),\ddot{std}).
% % \end{split}
% \end{equation}

\begin{equation}
% \begin{split}
\label{eq:loss_SDE}
L_{SDE}=\mathbb{D}(SDE(\ddot{y}),\ddot{std}).
% \end{split}
\end{equation}

\input{table/arch_SE}

Subsequently, we train the whole $SE$ in an end-to-end matter, where the $SDE$ and $TM$ modules in the $SE$ are trained using the following loss:
\begin{equation}
\begin{split}
\label{eq:SE_overall}
L_{SE}=\mathbb{D}(\hat{std^*},std^*)+\gamma_{std}\mathbb{D}(\hat{y}',\tilde{x}), 
\end{split}
\end{equation}
where $\mathbb{D}$ is the L1 distance, $\gamma_{std}$ is a weighting factor for balancing, $\hat{std^*}$ is the output of $SDE$, $\hat{y}'$ is the output of $TM$, and $\tilde{x} \equiv k^{std^*}_y\otimes x$. This allows the $SE$ to accurately predict the Gaussian blur level corresponding to any degraded input $y$.

Finally, as described in the main paper, the $SE$ and $RM$ are integrated to form the proposed DBLM, transforming the unknown degraded input $y$ into its corresponding Gaussian-blurred version $\acute{y}$. This transformation simplifies the Blind Face Restoration task into a Gaussian deblurring problem.

\subsection{Dynamic Starting Step Look-up Table}
As mentioned in Sec.4.2 in our main paper, the optimal starting timestep $t_{std}$ for each standard deviation is defined as follows:
\begin{equation}
\label{eq:SNR}
t_{std}=\underset{t}{argmin}\quad (log(\textbf{X}_{t})-log(\tilde{\textbf{Y}}_{t}^{std})\le tol),
\end{equation}
where $tol$ represents the maximum tolerance, $\textbf{X}_{t}$ and $\tilde{\textbf{Y}}_{t}^{std}$ denote the expected values of $x_{t}$ and $\tilde{y}_{t}^{std}$ in a training set.
We set $tol$ to $1 \times 10^{-3}$ to construct the Dynamic Starting Step Look-up Table (DSST), which pairs each standard deviation $std$ with its corresponding starting step $t_{std}$. Given the Gaussian-blur image $\acute{y}$, the estimated standard deviation $ \hat{std^*}$ from SE is equalized and serves as a key to retrieve the corresponding starting step from the DSST.
% , $std$ $\acute{y}$, the blur level of $\acute{y}$ is rounded to one decimal place to ensure precise matching within the DSST. This design ensures efficient and accurate retrieval of starting steps for guiding the diffusion model based on the blur level.

\begin{figure}[t]
    \centering
    \includegraphics[width=\linewidth]{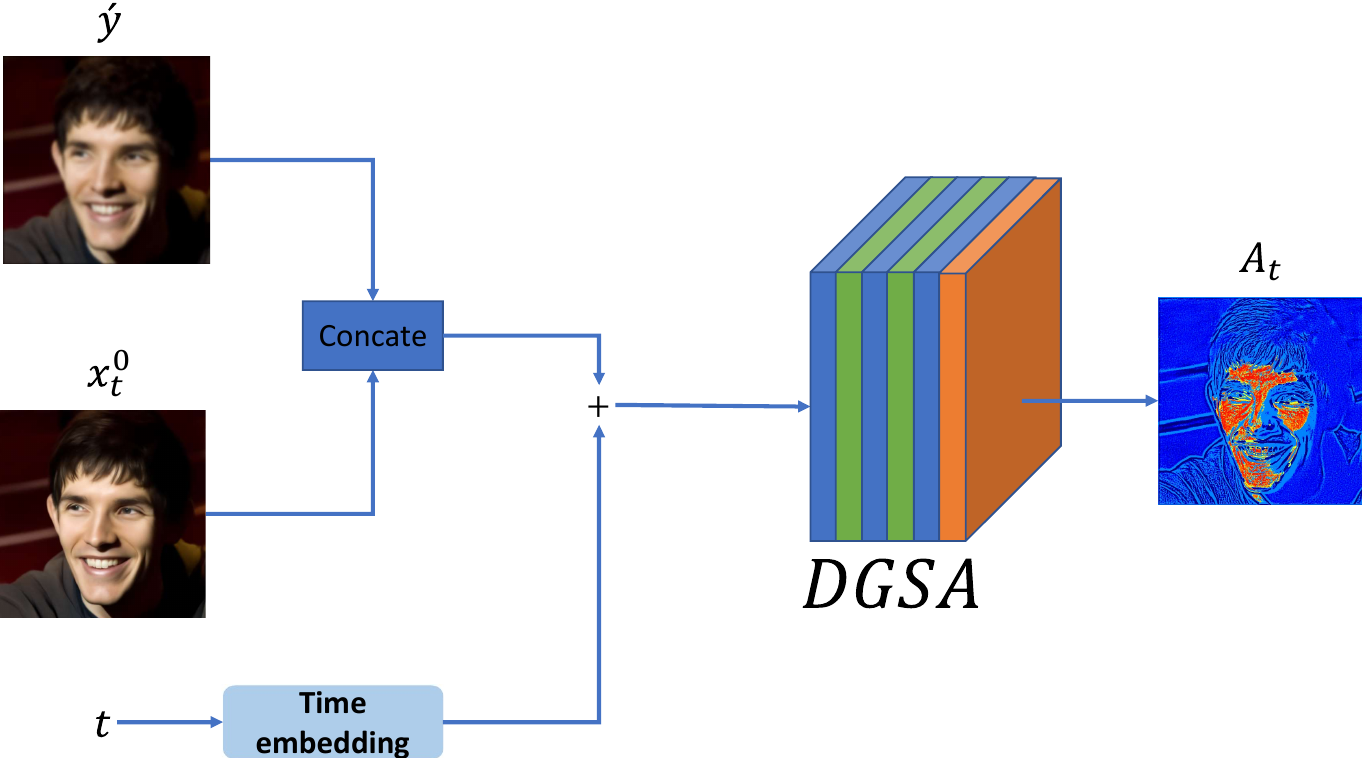}
    \caption{The inference flow of the DGSA is as follows: the measurement $\acute{y}$ and the high-quality prediction $x_t^0$ from $x_t$ are first concatenated to form a combined input. This concatenated input is then integrated with the current timestep $t$, which is processed through the time embedding module of the diffusion model. The resulting features are input into DGSA, generating a region-specific guidance scale map that dynamically adjusts the guidance scale at each timestep to balance fidelity and detail preservation.}
    \label{fig:DGSA}
\end{figure}

\input{table/arch_DGSA}

\subsection{Dynamic Guidance Scale Adjuster}

\noindent\textbf{Network Architecture:} The architecture of the Dynamic Guidance Scale Adjuster (DGSA) is outlined in \cref{tab:arch_DGSA}. DGSA consists of three convolutional layers, with ELU activation functions applied after the first two layers. The final layer uses the ReLU-1 activation function to constrain the output within the range of 0 to 1. At timestep $t$, the inputs to DGSA include the degraded measurement $\acute{y}$, the high-quality (HQ) prediction $x_t^0$ derived from $x_t$, and the current timestep $t$. 

In the implementation of DGSA, as illustrated in  \cref{fig:DGSA}, $\acute{y}$ and $x_t^0$ are concatenated before being input into the DGSA. This design ensures that the model effectively captures low-frequency information from both inputs to preserve fidelity. The timestep $t$ is then embedded and combined with the concatenated features to determine the localized diffusion power required at each region and timestep. The output of DGSA is a pixel-wise guidance scale map with values ranging from 0 to 1, matching the image dimensions. Higher values indicate stronger guidance influence to preserve fidelity, while lower values relax the guidance to utilize the DM’s realistic facial generation capabilities. This behavior is visualized in \cref{fig:dgsa_output}.

% The output of DGSA is a pixel-wise guidance scale map that matches the dimensions of $\acute{y}$ and $x_t^0$, enabling spatially localized adjustment of the guidance scale. This design ensures the diffusion model effectively balances detail generation with fidelity preservation across different image regions.

\input{algorithm/DGSA_training}

\noindent\textbf{Training procedure:} The DGSA training process is divided into two stages for optimal performance. Initially, DGSA is trained using actual Gaussian blurry images $\ddot{y}$ as the measurement inputs, generated based on \cref{eq:y_dot}. That is, we replace the output of DBLM $\acute{y}$ with $\ddot{y}$ in this stage. This stage consists of 20,000 iterations, allowing DGSA to learn robust guidance scale mappings for well-defined blurry images as its measurement input. In the second stage, DGSA undergoes fine-tuning with its original input, $\acute{y}$, representing the measurement predicted by our DBLM. This fine-tuning phase, spanning 7,000 iterations, further refines DGSA’s ability to adapt to real-world degraded inputs. During training, the DBLM module is kept frozen, and the DSST is pre-established.

We train DGSA at each randomly sampled timestep using the following loss function:
\begin{equation}
\begin{split}
    L_{DGSA}=  
    &\sum_{i}\gamma_{i}\mathbb{D}(SWT(x_{t-1}^{0})_{i},SWT(x_{0})_{i})
    + \\ &DISTS(x_{t-1}^{0},x_{0}),
\end{split}
\label{eq:L_DGSA}
\end{equation}
where {$\gamma_{i}$} are the weighted factors of the four subbands (LL, LH, HL, HH) decomposed by Stationary Wavelet Transformation ($SWT$) \cite{jawerth1994swt,korkmaz2024wgsr}, $x_0$ is the learning target, and $x_{t-1}^0$ is the HQ prediction based on $x_{t-1}$ at timestep $t-1$. Here, $\mathbb{D}$ and $DISTS$ \cite{ding2020dists,korkmaz2024wgsr} are the L1 reconstruction loss and the perceptual loss, respectively. 

The L1 reconstruction loss is applied across four subbands obtained via Stationary Wavelet Transform (SWT), with weighted factors $\gamma_{i}$ for LL, LH, HL, and HH set to 0.00, 0.01, 0.01, and 0.05, respectively. These weights prioritize high-frequency details, ensuring that the details of the samples adjusted by guidance remain consistent with the ground truth. Furthermore, we utilize DISTS to maintain the perceptual quality of the output. The complete training procedure is detailed in \cref{alg:DGSA}.

\subsection{Inference} 
Our DynFaceRestore framework is designed to be easily extendable, allowing for the use of multiple guidance sources. This flexibility enables a balance between perceptual quality and fidelity. In all our experiments, we utilize guidance from three blurred images, as outlined in \cref{alg:inference_mulguide}, with the weights $\lambda^{i\in[1,2,3]}$ are set to 0.7, 0.2, and 0.1, respectively. 
% based on the relative importance of low-frequency structural information and high-frequency details. % 
Further experimental analysis and discussions regarding the multiple guidance setup are provided in \cref{sec:mul_guide}.

\input{algorithm/inference_mulguide}

\section{Extra Experiments and Ablation Studies}
\label{sec:more_experiments}

\subsection{Kernel Mismatch}
\begin{figure}[t]
    \centering
    \includegraphics[width=\linewidth]{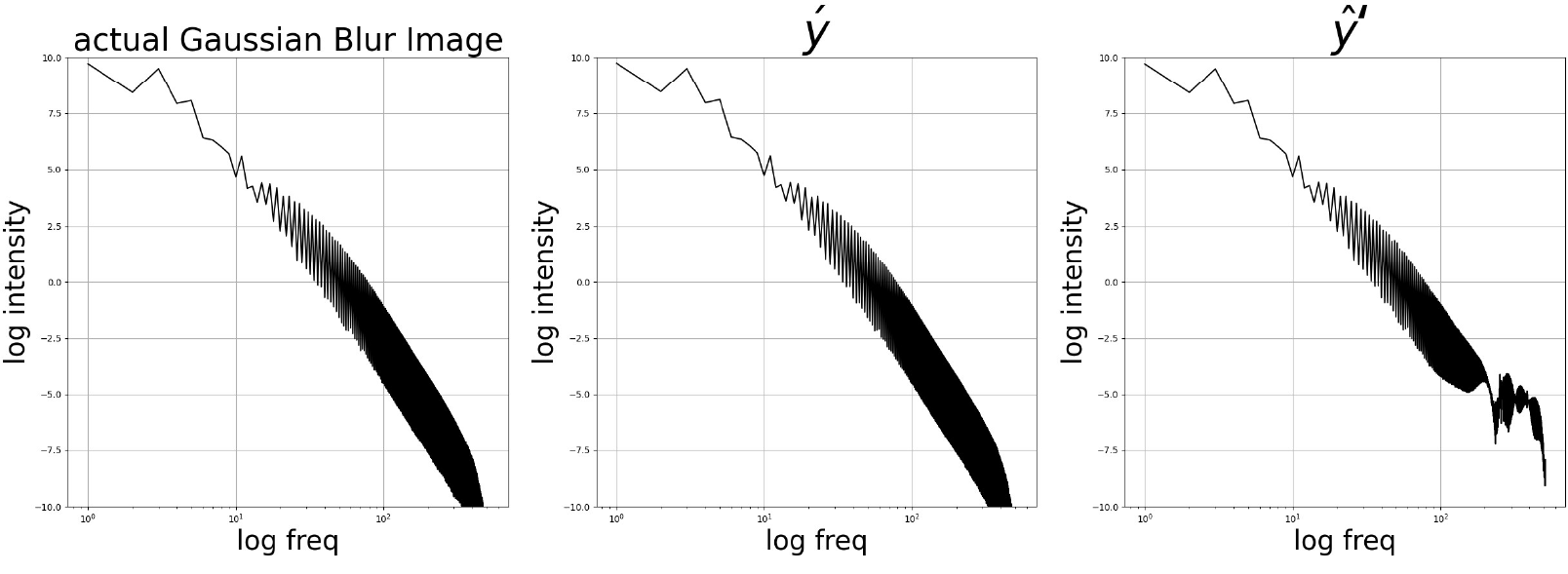}
    \caption{The frequency responses of actual Gaussian blurry images, $\acute{y}$, and $\hat{y}'$. Here, $\hat{y}'$, the output of $TM$ (refer to \cref{fig:SE}), exhibits significant differences in the high-frequency components compared to actual Gaussian-blurred images. In contrast, $\acute{y}$, estimated using \cref{eq:dblm_outputoutput}, closely matches the frequency response of the actual Gaussian-blurred image.}
    \label{fig:signal}
\end{figure}

\noindent\textbf{The difference between $\hat{y}'$ and $\acute{y}$:} The analysis in \cref{fig:signal} provides a detailed comparison between $\hat{y}'$ and $\acute{y}$, offering a strong rationale for designing the DBLM output as $\acute{y}$ rather than directly using the output of $TM$, $\hat{y}'$. By transforming $\hat{y}'$ and $\acute{y}$ into the frequency domain, a significant disparity in high-frequency intensity emerges when comparing $\hat{y}'$ to actual Gaussian-blurred images. 

In contrast, explicitly convolving the output of $RM(y)$ with a Gaussian kernel to produce $\acute{y}$ yields high-frequency intensity closely aligned with that of actual Gaussian-blurred images. This alignment suggests that the high-frequency discrepancy in $\hat{y}'$ is a crucial contributor to kernel mismatch, adversely affecting restoration fidelity. 
\CCH{
Furthermore, the quantitative results in \cref{tab:DBLM_output} confirm these findings, showing that using $\acute{y}$ as diffusion guidance in DBLM achieves superior fidelity over $\hat{y}'$, further validating the effectiveness of explicitly constructing $\acute{y}$.
}

\input{table/DBLM_output}

\begin{figure*}[th]
    \centering
    \includegraphics[width=\linewidth]{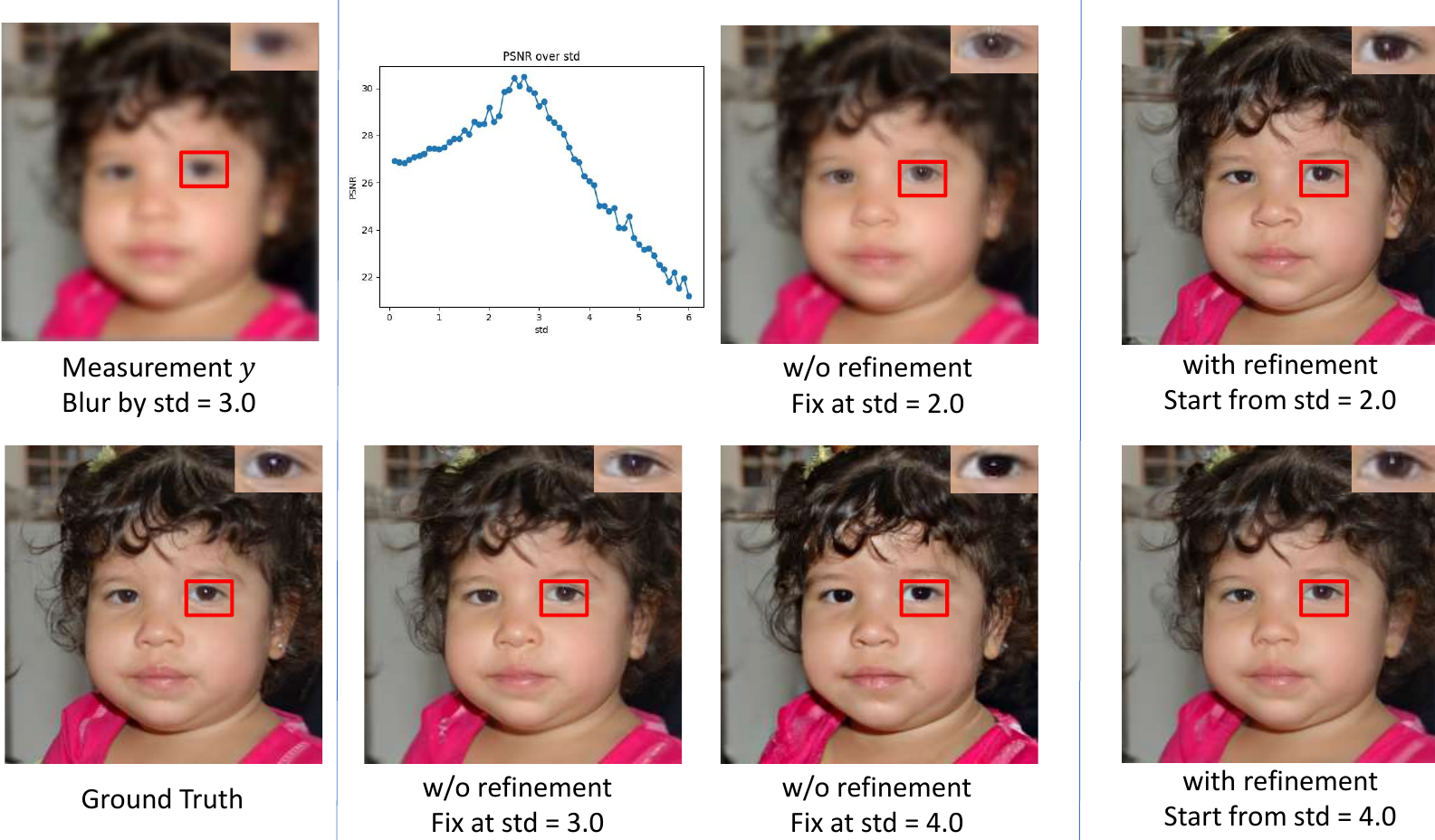}
    \caption{Comparison of results with and without kernel refinement is conducted using an actual Gaussian-blurred image with a kernel standard deviation (std) of 3.0 as the input measurement for guidance. During the diffusion sampling process, applying guidance adjustment without kernel refinement—e.g., fixing the kernel std to 2.0 or 4.0—leads to significant deviations from the ground truth due to kernel mismatch. When the fixed std is set to 2.0, the mismatch limits the diffusion model’s ability to add details, resulting in overly blurry outputs that rely too heavily on the measurement. Conversely, with a fixed std of 4.0, the diffusion model overly enhances the image, introducing hallucinated artifacts. In contrast, incorporating kernel refinement alongside guidance adjustment enables the sampling process to correct initial inaccuracies in kernel prediction. Even when starting with an imperfect kernel std, such as 2.0 or 4.0, the refined approach ensures that the final outputs align much more closely with the ground truth. This demonstrates the robustness of the refinement strategy in mitigating kernel mismatch issues.}
    \label{fig:kernel_refinement}
\end{figure*}

\noindent\textbf{Refinement of the Standard Deviation:} In \cref{fig:kernel_refinement}, experimental evidence highlights the critical role of kernel refinement at each step in mitigating kernel mismatch during the sampling process. Using an actual Gaussian blur measurement with a kernel of $std=3.0$ as the guidance, the experiment in \cref{fig:kernel_refinement} evaluates outcomes without employing DBLM, DSST, or DGSA to focus solely on analyzing the kernel refinement function. When the kernel std is not refined and fixed to the wrong standard deviation at each step, a substantial deviation between the final sampled results and the ground truth is evident, even with guidance applied. Notably, the results closely align with the ground truth only when the kernel std is accurately set to 3.0, as further validated by the ``PSNR over std'' analysis shown in \cref{fig:kernel_refinement}.

In contrast, with kernel std refinement applied at each step, the sampled results exhibit remarkable fidelity to the ground truth, even when the initial kernel std estimation deviates from the actual value. This outcome underscores the robustness of the kernel refinement strategy in addressing inaccuracies in initial kernel estimation, effectively bridging the gap caused by kernel mismatch. These findings validate the necessity of dynamic kernel std adjustments to achieve high-quality restoration.

\begin{figure}[th]
    \centering
    \includegraphics[width=\linewidth]{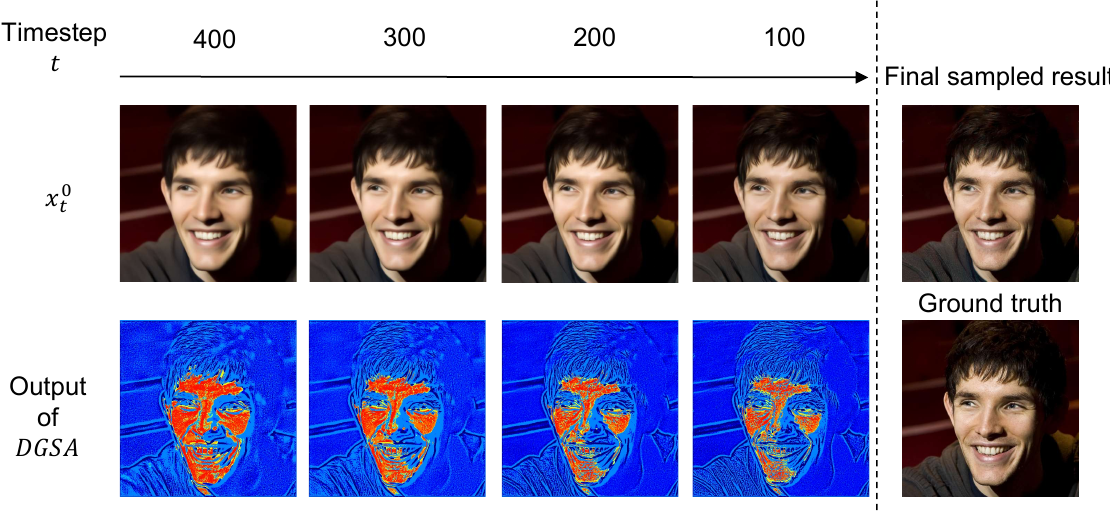}
    \caption{Visualization of DGSA at various timesteps during the sampling process. Here, $x_t^0$ is the HQ prediction of $x_t$ at timestep $t$. DGSA generates guidance maps, where blue regions indicate less guidance and rely more on the diffusion model (DM) to add details, while red regions signify stronger guidance with less reliance on the DM. As $t$ decreases, areas requiring the DM's HQ prior, such as hair, display a lower guidance scale to enhance detail. Conversely, facial regions maintain a higher guidance scale to ensure fidelity preservation.}
    \label{fig:dgsa_output}
\end{figure}

\subsection{Visualize of DGSA} The output of DGSA is visualized in \cref{fig:dgsa_output} to highlight its effectiveness. In each diffusion timestep, DGSA dynamically adjusts the guidance scale region-wise. The guidance scale gradually decreases for detail-rich areas such as hair and wrinkles, enabling these regions to harness more of the pre-trained DM’s high-quality image prior to sampling, resulting in realistic and refined facial details. Conversely, the guidance scale remains elevated for structural regions, such as the eyes and mouth, to preserve their geometric integrity and ensure the reconstructed image aligns closely with the ground truth. This adaptive balance between fidelity and detail generation leads to high realism outputs while maintaining structural consistency.

\subsection{Multiple Guidance}
\label{sec:mul_guide}
In this section, we evaluate the impact of varying the number of guidance sources. To ensure accurate assessment, DGSA is excluded to eliminate any potential external influences in this experiment. Each guidance source corresponds to a specific standard deviation, which changes following a defined pattern. For example, with \(n\) guidance sources,  we have \(\acute{y}'^{i \in [1, 2, \dots, n]}\) corresponding to the standard deviations \( \hat{std^*}, \hat{std^*}-1, \dots, \hat{std^*}-(n-1) \) and the weights $\lambda^{i \in [1, 2, \dots, n]}$.
 
Since $\acute{y}'^1$ with $\hat{std^*}$ retains more reliable low-frequency structural information, $\lambda^{1}$ is assigned a higher weight. In contrast, smaller standard deviations values  $\acute{y}'^{i\in[2,3,4]}$ provide higher-frequency details with reduced confidence, and thus, $\lambda^{i \in [2,3,4]}$ are assigned lower weights, as outlined by $\lambda^i$ in \cref{tab:mul_guide}. 

\input{table/mul_guide}

As shown in \cref{tab:mul_guide}, increasing the number of guidance sources amplifies the influence of the restoration model (RM), resulting in improved fidelity and higher scores in PSNR. However, perceptual quality (FID) is optimized when using a single guidance source, which allows more reliance on the diffusion model. This demonstrates the flexibility of our framework, enabling users to tune the desired output by controlling the number of guidance sources. Consequently, our framework effectively achieves a balance between perceptual quality and fidelity.

% As the result in \cref{tab:mul_guide}, increasing the number of guidance sources amplifies the influence of the restoration model (RM), thereby improving fidelity, yielding higher in both PSNR and IDA. However, this improvement comes at the cost of reduced diffusion model influence, resulting in diminished perceptual quality and higher FID score. This underscores the flexibility of our model, enabling the balancing of perceptual quality and fidelity through the management of guidance resources

\CCH{
\subsection{Different Starting Steps}
\label{sec:different_start_step}
By setting up the output of DBLM as a Gaussian-blurred image based on the input degradation severity, we can leverage this property to identify the optimal timestep ($t^*$), as outlined in Sec. 4.2 using ``Dynamic Starting Step Lookup Table''. Note that $t^*$ is automatically determined for different datasets. To demonstrate the effectiveness of our approach, we conduct experiments with different starting timesteps, as shown in \cref{tab:rebuttal_different_startingstep}. Larger timesteps ($>t^*$) result in a loss of information from the guidance observation, leading to reduced fidelity. Conversely, smaller timesteps ($<t^*$) provide insufficient iterations to recover fine details, thereby diminishing the quality of the reconstruction.

\input{table/rebuttal_different_step}

\subsection{Different SOTA Restoration Model}
\label{sec:different_RM}

\input{table/ablation_different_RM}

\begin{figure*}[t]
    \centering
    \includegraphics[width=0.6\linewidth]{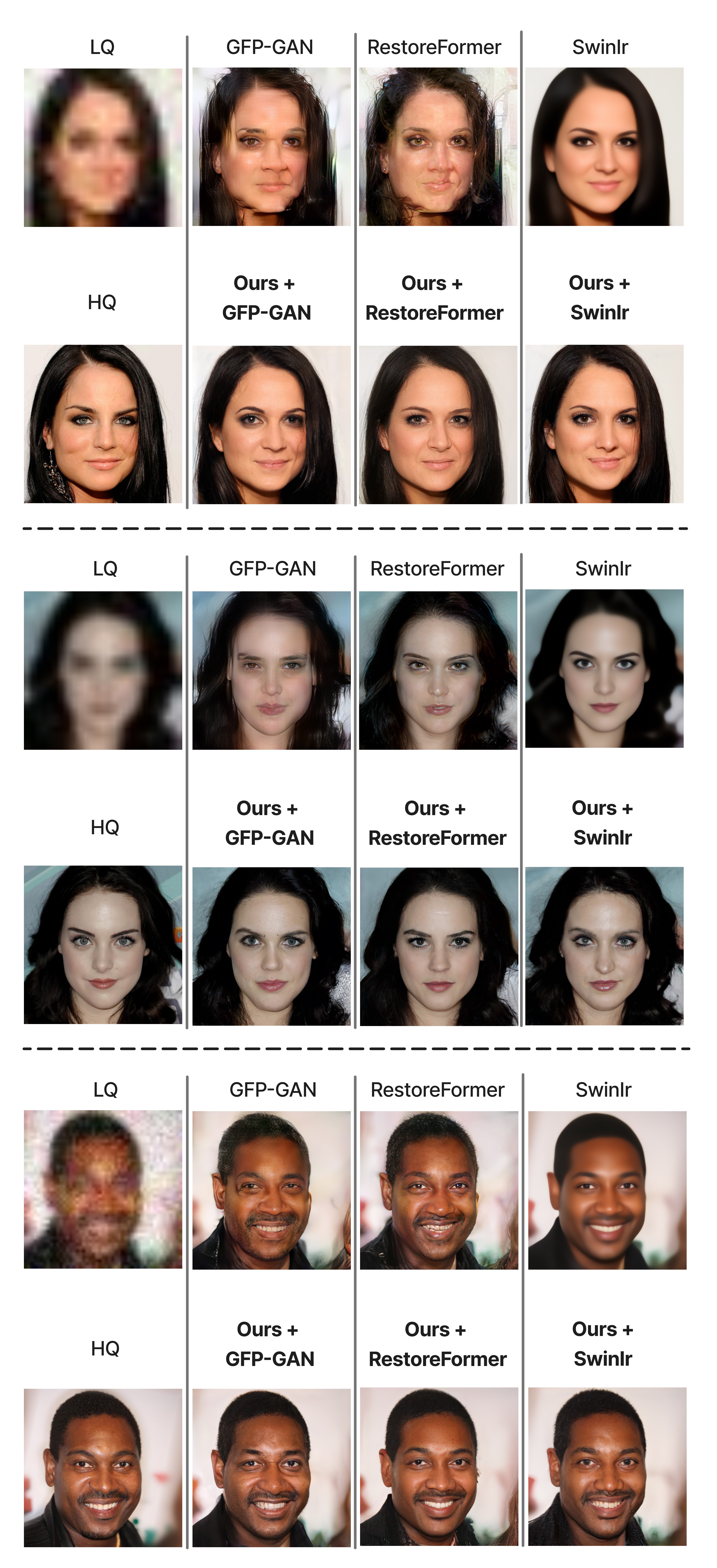}
    \caption{Ablation study with different RMs.}
    \label{fig:different_RM}
\end{figure*}

We evaluate the impact of different Restoration Models (RMs) on overall performance by replacing our pretrained RM with various alternatives, including GAN-based GFP-GAN \cite{kumar2022gfpgan}, CodeBook-based RestoreFormer \cite{wang2022restoreformer} and deterministic-based SwinIR \cite{liang2021swinir}. To ensure a fair comparison, we first fine-tune each RM on the FFHQ \cite{karras2019ffhq} dataset before integrating it into our framework.

As shown in \cref{tab:abaltion_differentRM} and visualized in \cref{fig:different_RM}, GFP-GAN suffers from poor realism (higher FID). Integrating our approach significantly improves the realism of GFP-GAN while maintaining competitive fidelity. Also, we find that RestoreFormer outperforms GFP-GAN in realism, and our method can further enhance RestoreFormer in terms of fidelity, structural integrity, PSNR and SSIM. Finally, SwinIR achieves the highest PSNR and SSIM but produces overly smooth images, compromising realism. Incorporating our method with SwinIR substantially reduces FID, balancing fidelity and realism. Overall, our method consistently improves realism across different RMs while maintaining competitive fidelity, achieving a more optimal trade-off between perceptual quality and fidelity.
}

\section{More Visualization}
\label{sec:more_visualization}
This section presents additional qualitative comparisons with state-of-the-art methods, including GPEN \cite{yang2021gpen}, GFPGAN \cite{kumar2022gfpgan}, RestoreFormer \cite{wang2022restoreformer}, CodeFormer \cite{zhou2022codeformer}, DiffBIR \cite{diffbir}, DAEFR \cite{tsai2024dualassociatedencoderface}, PGDiff \cite{yang2024pgdiff},  DifFace \cite{yue2024difface}, DR2 \cite{dr2}and 3Diffusion \cite{3Diffusion}.

\begin{figure*}[!htbp]
    \centering
    \includegraphics[width=0.825\linewidth]{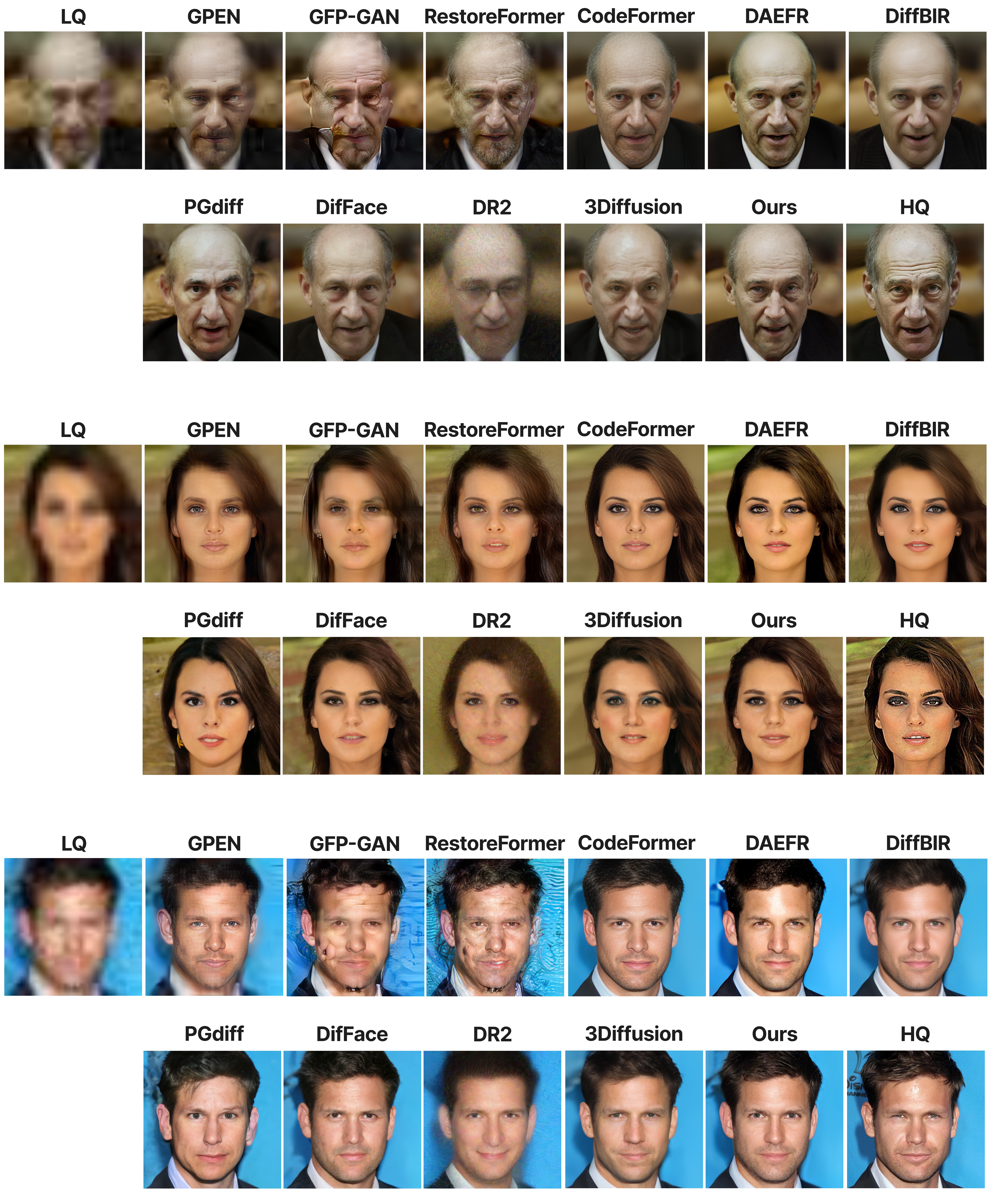}
    \caption{More visual comparisons on CelebA-Test. Our method achieves high-fidelity reconstructions while preserving natural facial features.}
    \label{fig:CelebA-Test_addition}
\end{figure*}

For the CelebA-Test dataset, \cref{fig:CelebA-Test_addition} demonstrates that our DynFaceRestore not only produces reconstructions closer to the ground truth than competing methods but also offers superior perceptual quality. This highlights the balance between fidelity and perceptual quality in our approach. 

\begin{figure*}[!htbp]
    \centering
    \includegraphics[width=0.825\linewidth]{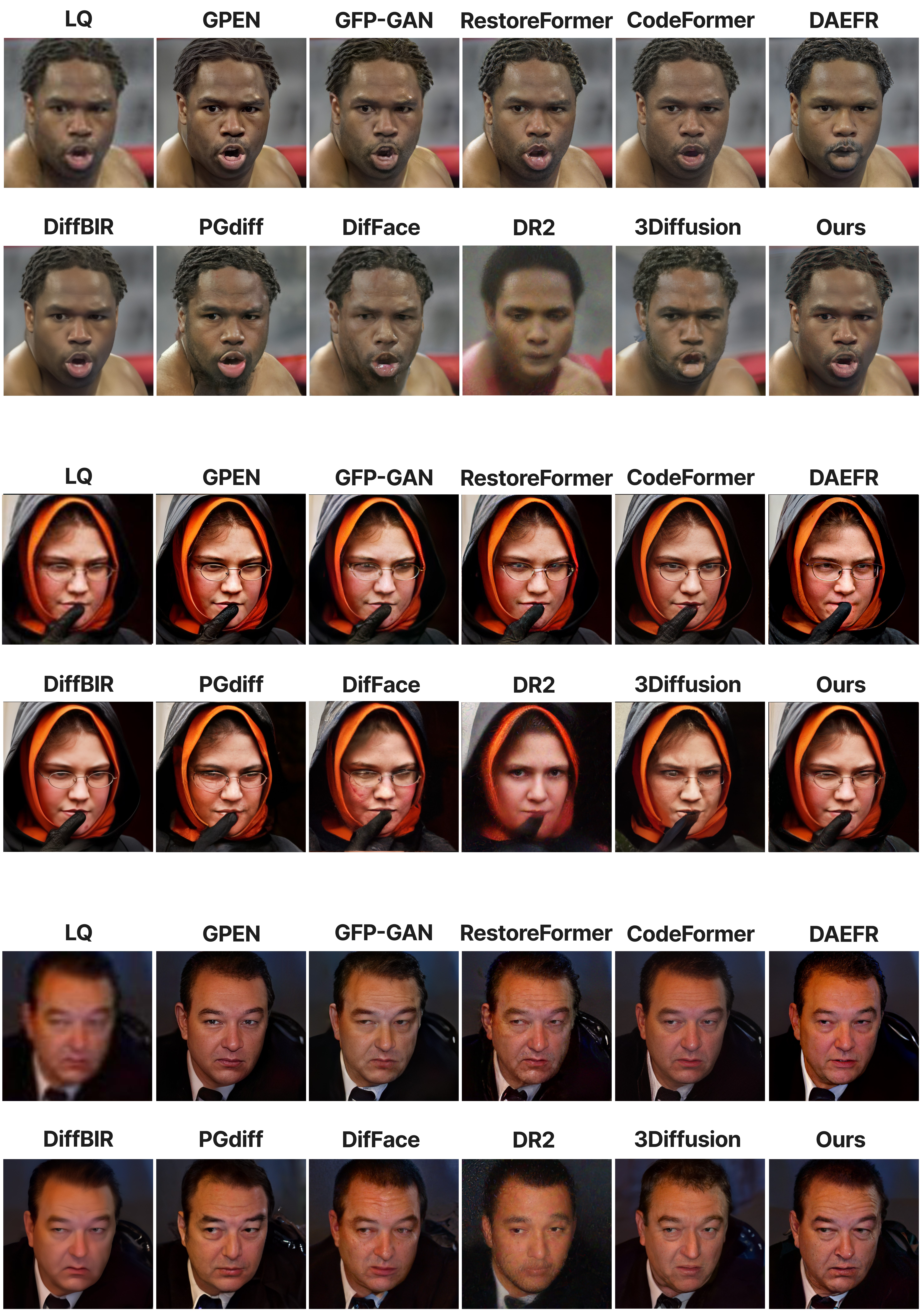}
    \caption{Qualitative results from LFW-Test demonstrate that our restoration method produces more natural features (e.g., eyes) and realistic details (e.g., hair) compared to other approaches, with improved fidelity.}
    \label{fig:LFW-Test_addition}
\end{figure*}

\begin{figure*}[!htbp]
    \centering
    \includegraphics[width=0.825\linewidth]{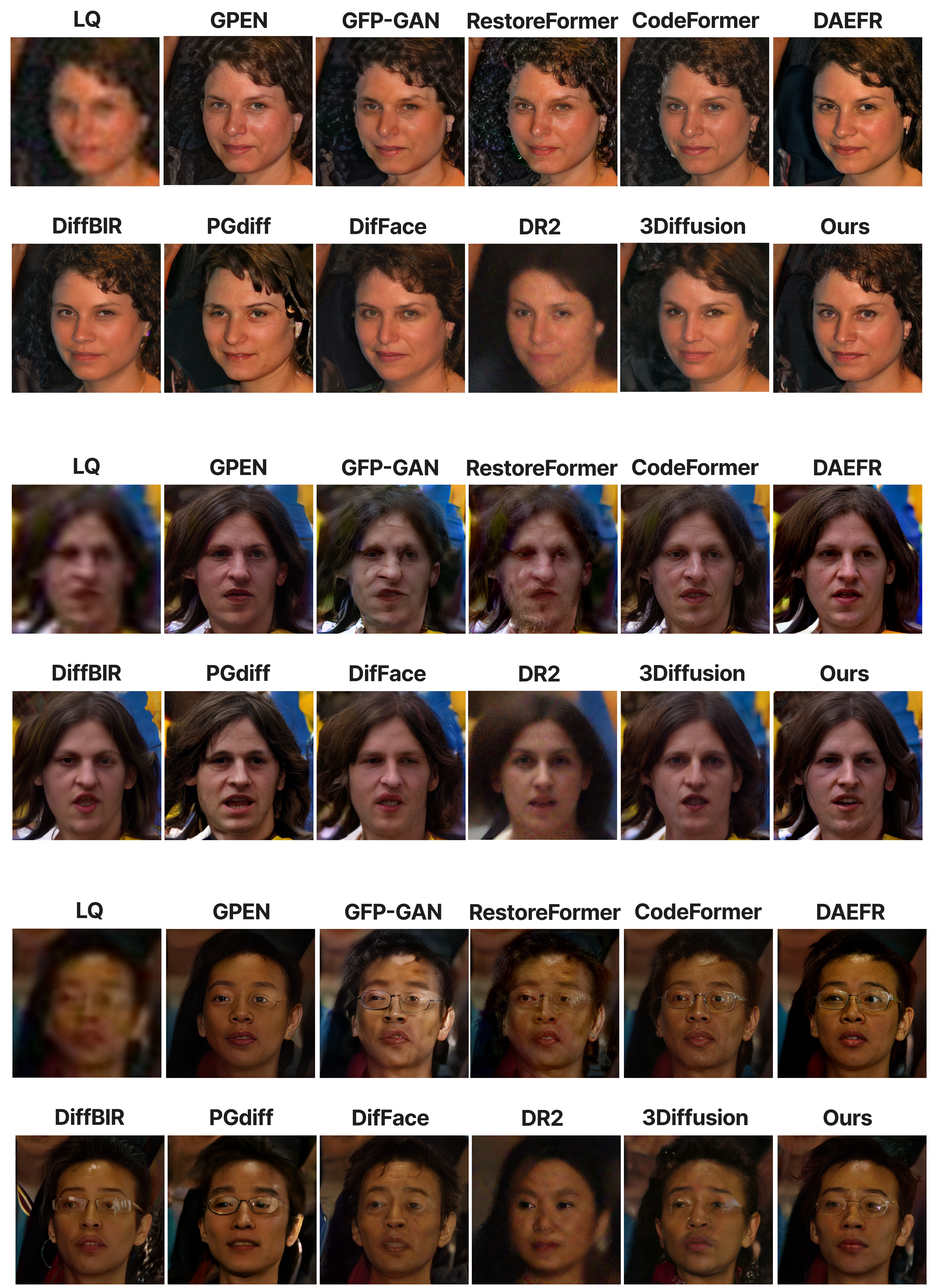}
    \caption{More visual comparisons on Wider-Test. Our restoration method produces more natural features (e.g., eyes) and realistic details (e.g., hair, skin) compared to other approaches, with improved fidelity.}
    \label{fig:Wider-Test_addition}
\end{figure*}

\begin{figure*}[!htbp]
    \centering
    \includegraphics[width=0.825\linewidth]{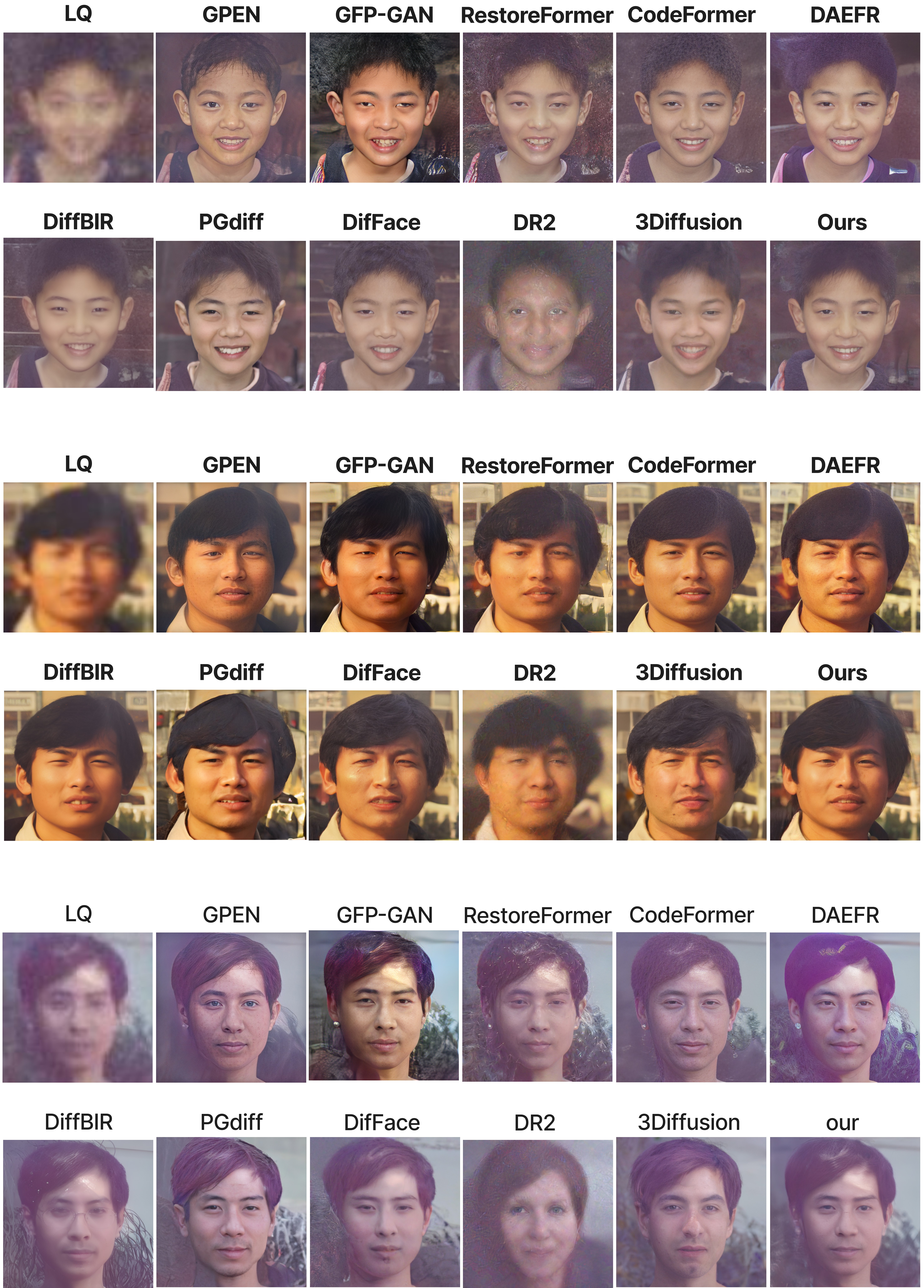}
    \caption{More visual comparisons on Webphoto-Test. Our restoration method produces more natural features (e.g., eyes) and realistic details (e.g., hair, skin) compared to other approaches, with improved fidelity.}
    \label{fig:Webphoto-Test_addition}
\end{figure*}

For the three real-world datasets—LFW-Test \cite{huang2008lfw}, Wider-Test \cite{zhou2022codeformer}, and Webphoto-Test \cite{kumar2022gfpgan}—the results in \cref{fig:LFW-Test_addition}, \cref{fig:Wider-Test_addition}, and \cref{fig:Webphoto-Test_addition}, respectively, highlight the advantages of DynFaceRestore. Specifically, our approach consistently generates more realistic facial details, such as hair strands and beards, while effectively preserving the global structure and local textures.

\begin{figure*}[h]
    \centering
    \includegraphics[width=1.0\linewidth]{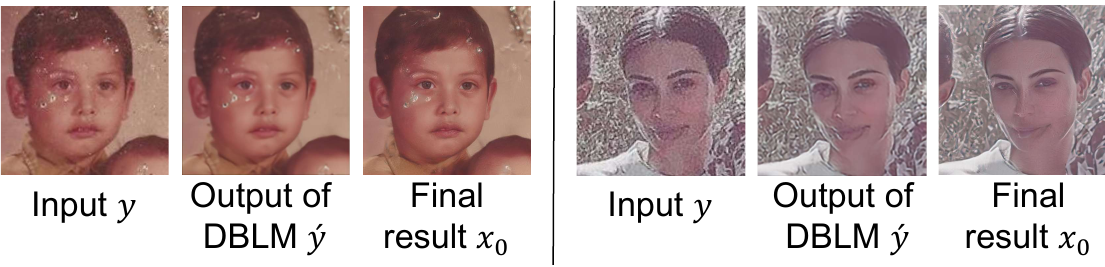}
    \caption{Limitations of Our DynFaceRestore. This issue likely arises from limitations in the degradation pipeline used to synthesize low-quality images for simulating real-world degradation. The current pipeline inadequately captures old photographs' diverse and severe degradation characteristics. Revising the pipeline to represent these complexities better is essential for improving restoration performance in such challenging scenarios.}
    \label{fig:limitations}
\end{figure*}

\begin{figure}[!h]
    \centering
    \includegraphics[width=0.9\linewidth]{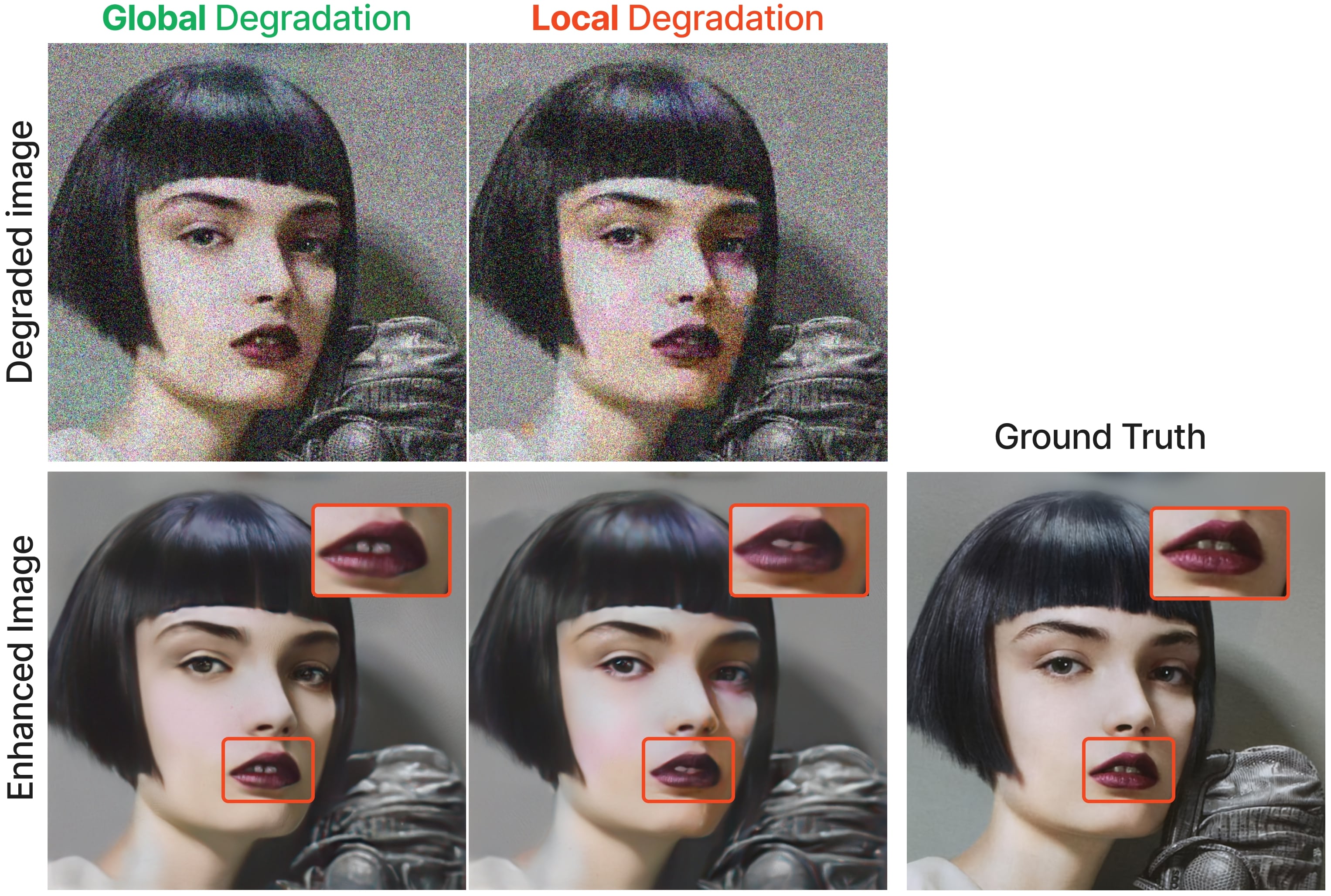}
    \caption{Limitations of Our DynFaceRestore. We degrade images using \cref{eq:degradation_pipeline}, applying both global degradation (uniformly across the entire image) and local degradation (independently for different small regions). The results indicate the appearance of artifacts in the mouth and eye regions.}
    \label{fig:limitations_2}
\end{figure}

\section{Limitations}
\label{sec:limitations}
\CCH{
A major limitation of our proposed method is its computational complexity, as shown in Tab. 1 (main paper). This is primarily due to the optimization of kernel prediction and the incorporation of DM guidance at each step. The acceleration through DDIM \cite{ddim} has yet to be explored. Therefore, reducing the denoising steps and selectively applying guidance to the critical step offers a promising research direction to address this challenge.

Additionally, as illustrated in \cref{fig:limitations}, the DBLM module in our DynFaceRestore framework exhibits limitations when handling the complex degradation patterns observed in old photographs, such as incomplete restoration due to residual artifacts (\cref{fig:limitations} left) or geometric distortions (\cref{fig:limitations} right). These limitations result in difficulties during the subsequent DM sampling and guidance processes, leading to difficulties in accurately identifying regions that require fidelity preservation and refinement. As a result, the overall restoration quality is degraded. We attribute this issue to the training of the pre-trained restoration model, which fails to capture the diverse and severe degradations commonly observed in old photographs. A straightforward and effective solution to overcome these limitations is to replace the RM with more advanced restoration models. 
% Additionally, training the RM using unsupervised or self-supervised learning paradigms has the potential to better capture the diverse and severe degradations typically encountered in real-world scenarios, such as those present in old photographs. 

Moreover, real-world images often exhibit spatially varying degradations, posing a significant challenge in perfectly addressing the kernel mismatch issue. Treating the entire image as uniformly degraded can lead to undesired artifacts, as shown in \cref{fig:limitations_2}, which our proposed method has not yet effectively resolved. Addressing this issue is crucial for future research, and a direct approach is to divide regions and estimate degradations separately. 
% Extending our current work to address these issues would be a direction for future research.
}

%% file: table/arch_SE.tex
\begin{table}
\small
\centering
\caption{Architecture of $SDE$.}
\label{tab:arch_se}
\begin{tabular}{|c|c|}
    \hline
    architecture & channels \\ 
    \hline
    Conv2d: kernel size: 3$\times$3; stride: 1 & 3 $\rightarrow$ 64\\ 
    %\hline
    BatchNormalize2d & 64\\ 
    %\hline
    LeakyReLU & - \\
    \hline
    Conv2d: kernel size: 3$\times$3; stride: 1 & 64 $\rightarrow$ 64\\ 
    %\hline
    BatchNormalize2d & 64\\ 
    %\hline
    LeakyReLU & - \\
    \hline
    Conv2d: kernel size: 3$\times$3; stride: 2 & 64 $\rightarrow$ 128\\ 
    %\hline
    BatchNormalize2d & 128\\ 
    %\hline
    LeakyReLU & - \\
    \hline
    Conv2d: kernel size: 3$\times$3; stride: 1 & 128 $\rightarrow$ 128\\ 
    %\hline
    BatchNormalize2d & 128\\ 
    %\hline
    LeakyReLU & - \\
    \hline
    Conv2d: kernel size: 3$\times$3; stride: 2 & 128 $\rightarrow$ 256\\ 
    %\hline
    BatchNormalize2d & 256\\ 
    %\hline
    LeakyReLU & - \\
    \hline
    Conv2d: kernel size: 3$\times$3; stride: 1 & 256 $\rightarrow$ 256\\ 
    %\hline
    BatchNormalize2d & 64\\ 
    %\hline
    LeakyReLU & - \\
    \hline
    AvgPool2d & - \\
    \hline
    Linear & 256 $\rightarrow$ 256\\
    LeakyReLU & - \\
    Linear & 256 $\rightarrow$ 256\\
    Linear & 256 $\rightarrow$ 1\\
    \hline
\end{tabular}
\end{table}

%% file: table/arch_DGSA.tex
\begin{table}
\small
\centering
\caption{Network architecture of DGSA.}
\label{tab:arch_DGSA}
\begin{tabular}{|c|c|}
    \hline
    architecture & channels \\ 
    \hline
    Conv2d: kernel size: 3$\times$3 & 6 $\rightarrow$ 64 \\
    ELU & - \\
    \hline
    Conv2d: kernel size: 3$\times$3 & 64 $\rightarrow$ 64 \\
    ELU & - \\
    \hline
    Conv2d: kernel size: 3$\times$3 & 64 $\rightarrow$ 3 \\
    ReLU-1 & - \\
    \hline
\end{tabular}
\end{table}

%% file: algorithm/DGSA_training.tex
\begin{algorithm}[th]
\small
  \caption{DGSA training}
  \label{alg:DGSA}
  \begin{algorithmic}[1]   
    \Require
      $y$: Unknown degraded LQ input;
      $iter_{total}$: Total training iterations;
      $\gamma_{i}$: Weight factor of SWT four subbands;
    \State $iter$ = 0;
    \While {$iter<iter_{total}$}
    \State $\acute{y},\hat{std^*}=DBLM(y), SE(y)$;
    \State $t_{start}=DSST(\hat{std^*})$;
    \State $t \sim Uniform(0, t_{start})$;
    \State $x_{t}=\sqrt{\bar{\alpha}_{t}}x_{0}+\sqrt{1-\bar{\alpha}_{t}}\epsilon,\space\epsilon\sim N(0,1)$;
    \State $x_{t}^{0}=\frac{1}{\sqrt{\bar{\alpha_{t}}}}x_{t}-\sqrt{\frac{1-\bar{\alpha_{t}}}{\bar{\alpha}_{t}}}\epsilon_{\theta}$;
    \State compute Gaussian blur kernel $k$ using $\hat{std^*}$;
    \State $x_{t-1}'=\frac{1}{\sqrt{\alpha_{t}}}(x_{t}-\frac{\beta_{t}}{\sqrt{1-\bar{\alpha}_{t}}}\epsilon_{\theta})+\sigma_{t}\epsilon,\space\epsilon\sim N(0,1)$;
    \State $A_t=DGSA(\acute{y},x_{t}^{0},t)$
    \State $x_{t-1}=x_{t-1}'-A_t\times \nabla_{x_{t}}\left\| \acute{y}-k_{t}\otimes x_{t}^{0} \right\|^{2}$;
    \State $x_{t-1}^{0}=\frac{1}{\sqrt{\bar{\alpha_{t-1}}}}x_{t-1}-\sqrt{\frac{1-\bar{\alpha_{t-1}}}{\bar{\alpha}_{t-1}}}\epsilon_{\theta}$;
    \State $L_{DGSE}=  
    \sum_{i}\gamma_{i}\mathbb{D}(SWT(x_{t-1}^{0})_i,SWT(x_{0})_i)$\\
    $\quad\quad\quad\quad\quad\quad+\quad DISTS(x_{t-1}^{0},x_{0})$;
    \State update DGSA using $L_{DGSE}$;
    \State $iter = iter + 1$;
    \EndWhile
  \end{algorithmic}
\end{algorithm}

%% file: algorithm/inference_mulguide.tex
\begin{algorithm*}[h]
% \small
  \caption{Inference - Multiple Guidance}
  \label{alg:inference_mulguide}
  \begin{algorithmic}[1]
    \Require
      $y$: Unknown degraded LQ input; 
      $\lambda^{i\in[1,2,3]}$: weights of each guidance;
    \Ensure
      $x_{0}$: HQ sampled image;    
    \State $\acute{y},\hat{std^*}=DBLM(y), SE(y)$;
    \State $\hat{std^*}^{i\in[1,2,3]} = [\hat{std^*},\hat{std^*}-1,\hat{std^*}-2]$
    \State compute Gaussian blur kernel $k^{i\in[1,2,3]}$ using $\hat{std^*}^{i\in[1,2,3]}$;
    \State $\acute{y}^{i\in[1,2,3]}=[\acute{y},k^{2}\otimes RM(y),k^{3}\otimes RM(y)]$
    \State $t_{start}^{i\in[1,2,3]}=DSST(\hat{std^*}^{i\in[1,2,3]})$;
    \State $std_{t_{start}}^{i\in[1,2,3]}=\hat{std^*}^{i\in[1,2,3]}$;
    \State $x_{t_{start}}=\sqrt{\bar{\alpha}_{t_{start}^{1}}}\acute{y}^{1}+\sqrt{1-\bar{\alpha}_{t_{start}^{1}}}\epsilon, \epsilon\sim N(0,1)$;
    \For{$t=t_{start}^{1}\cdots 1$}
      \State $x_{t}^{0}=\frac{1}{\sqrt{\bar{\alpha_{t}}}}x_{t}-\sqrt{\frac{1-\bar{\alpha_{t}}}{\bar{\alpha}_{t}}}\epsilon_{\theta}$;
      \State $x_{t-1}=\frac{1}{\sqrt{\alpha_{t}}}(x_{t}-\frac{\beta_{t}}{\sqrt{1-\bar{\alpha}_{t}}}\epsilon_{\theta})+\sigma_{t}\epsilon,\space\epsilon\sim N(0,1)$;
      \State compute Gaussian blur kernel $k_{t}^{i\in[1,2,3]}$ using $std_{t}^{i\in[1,2,3]}$;
      \State $A_t=DGSA(\acute{y}^1,x_{t}^{0},t)$;
      \If{$t\in[t_{start}^1,t_{start}^2]$}
        \State $x_{t-1}=x_{t-1}'-A_t\times 
\nabla_{x_{t}}\left\| \acute{y}^{1}-k_{t}^{1}\otimes x_{t}^{0} \right\|^{2}$;
        \State $std_{t-1}^{1}=std_{t}^{1}-\sqrt{\bar{\alpha}_{t}}\times 
\nabla_{k_{t}^{1}}\left\| \acute{y}^1-k_{t}^1\otimes x_{t}^{0} \right\|^{2}$;
      \EndIf
      \If{$t\in[t_{start}^2,t_{start}^3]$}
        \State $x_{t-1}=x_{t-1}'-A_t\times 
\nabla_{x_{t}}\sum_{i=1}^{2}(\frac{\lambda^{i}}{\lambda^{1}+\lambda^{2}})\left\| \acute{y}^{i}-k_{t}^{i}\otimes x_{t}^{0} \right\|^{2}$;
        \State $std_{t-1}^{i\in[1,2]}=std_{t}^{i\in[1,2]}-\sqrt{\bar{\alpha}_{t}}\times 
\nabla_{k_{t}^{i\in[1,2]}}(\frac{\lambda^{i}}{\lambda^{1}+\lambda^{2}})\left\| \acute{y}^{i\in[1,2]}-k_{t}^{i\in[1,2]}\otimes x_{t}^{0} \right\|^{2}$;
      \EndIf
      \If{$t\in[t_{start}^3,0]$}
        \State $x_{t-1}=x_{t-1}'-A_t\times 
\nabla_{x_{t}}\sum_{i=1}^{3}(\frac{\lambda^{i}}{\lambda^{1}+\lambda^{2}+\lambda^{3}})\left\| \acute{y}^{i}-k_{t}^{i}\otimes x_{t}^{0} \right\|^{2}$;
        \State $std_{t-1}^{i\in[1,2,3]}=std_{t}^{i\in[1,2,3]}-\sqrt{\bar{\alpha}_{t}}\times 
\nabla_{k_{t}^{i\in[1,2,3]}}(\frac{\lambda^{i}}{\lambda^{1}+\lambda^{2}+\lambda^{3}})\left\| \acute{y}^{i\in[1,2,3]}-k_{t}^{i\in[1,2,3]}\otimes x_{t}^{0} \right\|^{2}$;
      \EndIf
    \EndFor \\
  \Return $x_{0}$
  \end{algorithmic}
\end{algorithm*}

%% file: table/DBLM_output.tex
\begin{table}[!h]
\small
\centering
\caption{Ablation study on different output types of DBLM in the CelebA-Test dataset. Here, $\hat{y}'$ and $\acute{y}$ represent different approximations of the Gaussian-blurred image $\tilde{x}$. The results clearly show that $\acute{y}$ provides a closer approximation of $\tilde{x}$, resulting in improved fidelity.}
%The best performance is highlighted with {\bf bold}.}
\label{tab:DBLM_output}
% \begin{tabular}{|c|c|c|}
%     \hline
%     Output type of DBLM & PSNR$\uparrow$ & SSIM$\uparrow$ \\ 
%     \hline
%      $\hat{y}'$ & 24.915 & 0.684 \\
%     %\hline
%      $\acute{y}$ & \bf24.992 & \bf0.690 \\
%     \hline
% \end{tabular}
% \end{table}

\scalebox{0.8}{
\begin{tabular}{|c|c|c|c|c|c|c|}
    \hline
    \makecell{Gaussian Blur\\image}  & PSNR$\uparrow$ & SSIM$\uparrow$ & LPIPS$\downarrow$ & FID$\downarrow$ & IDA$\downarrow$ & LMD$\downarrow$ \\ 
    \hline
     $\hat{y}'$ & 24.261 & 0.661 & 0.332 & \bf14.185 & 0.756 & 3.428 \\
    %\hline
     $\acute{y}$ & \bf24.349 & \bf0.664 & 0.332& 14.780 & \bf0.748 & \bf3.419  \\
    \hline
\end{tabular}
}
\end{table}

%% file: table/mul_guide.tex
\begin{table}
\small
\centering
\caption{Ablations of the different numbers of guidance in CelebA-Test dataset. $\#$ denotes the numbers of guidance and $\lambda^i$ are the weights of each guidance. The best performance is highlighted with {\bf bold}.}
\label{tab:mul_guide}
\begin{tabular}{|c|c|c|c|c|}
    \hline
    $\#$ & $\lambda^i$ & PSNR$\uparrow$ & IDA$\downarrow$ & FID$\downarrow$ \\ 
    \hline
     1 & [1.0] & 25.014 & 0.7242 & \bf18.452 \\
    %\hline
     2 & [0.8, 0.2] & 25.049 & 0.7238 & 18.98 \\
    % \hline
     \rowcolor{Gray}3 & [0.7, 0.2, 0.1] & 25.107 & \bf0.7236 & 19.786 \\
    % \hline
     4 & [0.7, 0.1, 0.1, 0.1] & \bf25.185 & 0.7242 & 20.947 \\
    \hline
\end{tabular}
\end{table}

%% file: table/rebuttal_different_step.tex
\begin{table}[!h]
\small
% \captionsetup{skip=0mm}
% \centering
\caption{Ablation study of same blurred image with different starting steps in CelebA-Test. Top performances in {\bf bold} and \underline{underline}.}
% Comparisons of different starting steps in CelebA-Test dataset. The best and second performances are highlighted with {\bf bold} and \underline{underline}.
\label{tab:rebuttal_different_startingstep}
\scalebox{0.82}{
\begin{tabular}{|c|c|c|c|c|c|c|}
    \hline
    Steps & PSNR$\uparrow$ & SSIM$\uparrow$ & LPIPS$\downarrow$ & FID$\downarrow$ & IDA$\downarrow$ & LMD$\downarrow$ \\ 
    \hline
    400 ($<t^*$) & 23.393 & 0.659 & 0.394 & 42.055 & 0.853 & 3.781 \\
    % \hline
    1000 ($>t^*$) & \underline{24.172} & \underline{0.657} & \underline{0.336} & \bf14.556 & \underline{0.761} & \underline{3.470} \\
    % \hline
    \rowcolor{Gray}$t^*$  [690,925] & \bf24.349 & \bf0.664 & \bf{0.332} & \underline{14.780} & \bf0.748 & \bf3.419 \\
    \hline
    % & \rowcolor{Gray} DynFaceRestore & \bf24.047 & \underline{0.653} & \underline{0.340} & \bf14.539 & 4.156 & 0.783 & \underline{3.615} \\
    % \hline
\end{tabular}

}
\vspace{-3mm}
\end{table}

%% file: table/ablation_different_RM.tex
\begin{table*}[!h]
\small
\centering
\caption{Differnet RM model comparisons to in CelebA-Test. The best and second performances are highlighted with {\bf bold} and \underline{underline}.}
\label{tab:abaltion_differentRM}
\scalebox{1}{
% \begin{tabular}{|c|c|c|c|c|c|c|c|lll}
\begin{tabular}{|c|c|c|c|c|c|c|c|}

    \hline
    % Type & Method & \makecell{Inference\\time (s)} & \makecell{MACs\\ (G)}  & \makecell{Params\\ (M)} &PSNR$\uparrow$ & SSIM$\uparrow$ & LPIPS$\downarrow$ & FID$\downarrow$ & IDA$\downarrow$ & LMD$\downarrow$  \\ 
    Type & Method &PSNR$\uparrow$ & SSIM$\uparrow$ & LPIPS$\downarrow$ & FID$\downarrow$ & IDA$\downarrow$ & LMD$\downarrow$  \\ 
    
    \hline
    
    \multirow{2}{*}{GAN}
    % & GPEN&23.773& \bf0.659& \bf0.358& 30.250& 0.837& 6.377\\
    %\hline
                                    % & Ours + GPEN&22.973& 0.644& 0.364& \bf20.847& \bf0.837& \bf4.407\\
       % \cline{2-8}                           
 & GFP-GAN& 22.841& 0.620& 0.355& 23.860& 0.822&4.793\\ 
   
                                     & Ours + GFP-GAN& \bf23.867& \bf0.649& \bf0.340& \bf15.943& \bf0.772& \bf3.654\\
    \hline
    \multirow{2}{*}{CodeBook}
    % & DAEFR&21.715& 0.591& 0.343& \bf15.827& \bf0.863& \bf3.848\\
                                     % & Ours + DAEFR&\bf23.241& \bf0.648& \bf0.353& 18.840& 0.905& 4.275\\
    % \cline{2-8}
                                     & RestoreFormer&23.001& 0.592& 0.376& 22.874& 0.783& 4.464\\
    %\hline
       & Ours + RestoreFormer&\bf23.794& \bf0.655& \bf0.341& \bf17.042& \bf0.820& \bf3.859\\
    \hline
    \multirow{2}{*}{Deterministic}& SwinIR&\bf26.177& \bf0.746& 0.377& 61.209& \bf0.720& \bf3.101\\ 
    %\hline
           &\cellcolor{Gray}  Ours + SwinIR & \cellcolor{Gray} 24.349& \cellcolor{Gray}  0.664& \cellcolor{Gray} 
 \bf0.332& \cellcolor{Gray}  \bf14.78&  \cellcolor{Gray} 0.748&  \cellcolor{Gray} 3.419\\
    \hline
    % & \rowcolor{Gray} DynFaceRestore & \bf24.047 & \underline{0.653} & \underline{0.340} & \bf14.539 & 4.156 & 0.783 & \underline{3.615} \\
    % \hline
\end{tabular}
}
\end{table*}

% \begin{table*}
% \small
% \centering
% \caption{Comparisons to SOTA methods in CelebA-Test. The best and second performances are highlighted with {\bf bold} and \underline{underline}.}
% \label{tab:celeba_mix}
% \begin{tabular}{|c|c|c|c|c|c|c|c|c|}
%     \hline
%     Type & Method & PSNR$\uparrow$ & SSIM$\uparrow$ & LPIPS$\downarrow$ & FID$\downarrow$ & NIQE$\downarrow$ & IDA$\downarrow$ & LMD$\downarrow$ \\ 
%     \hline
%     \multirow{2}{*}{GAN} & GPEN & 23.444 & 0.638 & 0.374 & 25.662 & 4.688 & 0.789 & 5.919 \\ 
%     %\hline
%         & GFP-GAN & 22.971 & 0.623 & 0.352 & 23.401 & 4.132 & 0.801 & 4.693 \\
%     \hline
%     \multirow{3}{*}{Codebook}  & RestoreFormer & 23.140 & 0.596 & 0.374 & 22.660 & 4.279 & 0.763 & 4.361 \\
%     %\hline
%         & CodeFormer & 23.938 & 0.639 & \bf0.317 & 17.759 & 4.630 & \underline{0.759} & \bf3.443 \\ 
%     %\hline
%         & DAEFR & 21.715 & 0.591 & 0.343 & 15.827 & \underline{4.036} & 0.863 & 3.848 \\
%     \hline
%     \multirow{3}{*}{pretrained DM} & PGDiff & 21.824 & 0.612 & 0.369 & 20.928 & \bf3.955 & 0.944 & 4.868 \\
%     %\hline
%        & DifFace & \underline{23.949} & \underline{0.659} & 0.355 & \underline{15.032} & 4.451 & 0.867 & 3.781 \\
%     %\hline
%        & \rowcolor{Gray} DynFaceRestore & \bf24.177 & \bf{0.660} & \underline{0.334} & \bf14.356 & 4.239 & \bf0.740 & \underline{3.479} \\
%     \hline
%     % & \rowcolor{Gray} DynFaceRestore & \bf24.047 & \underline{0.653} & \underline{0.340} & \bf14.539 & 4.156 & 0.783 & \underline{3.615} \\
%     % \hline
% \end{tabular}
% \end{table*}